\PassOptionsToPackage{table}{xcolor}
\PassOptionsToClass{balance=false}{acmart}
\def\HFSFullVersion{1}
 \documentclass[sigconf]{acmart}
\newif\ifHFSAppendix
\ifdefined\HFSFullVersion
  \HFSAppendixtrue
\else
  \HFSAppendixfalse
\fi
 \usepackage{bm}
 \definecolor{lightpurple}{RGB}{230, 230, 250}
 \usepackage{multirow}

\newenvironment{promptbox}[1]{%
  \par\addvspace{7pt}\noindent
  \colorbox{lightpurple}{\begin{minipage}{\dimexpr\columnwidth-2\fboxsep\relax}%
  \strut\textbf{#1}\strut\end{minipage}}%
  \par\nobreak\nointerlineskip\vskip4pt
  \nobreak\small\rmfamily\noindent\ignorespaces
}{%
  \par\nointerlineskip\vskip3pt
  {\color{lightpurple}\hrule height 2.5pt}\addvspace{7pt}\noindent
}

\AtBeginDocument{%
  }

\copyrightyear{2026}
\acmYear{2026}
\setcopyright{cc}
\setcctype{by}
\acmConference[MM '26]{Proceedings of the 34th ACM International Conference on Multimedia}{November 10--14, 2026}{Rio de Janeiro, Brazil}
\acmBooktitle{Proceedings of the 34th ACM International Conference on Multimedia (MM '26), November 10--14, 2026, Rio de Janeiro, Brazil}
\acmDOI{10.1145/3767308.3836416}
\acmISBN{979-8-4007-2213-4/2026/11}
\acmSubmissionID{mfp8437}

\begin{document}

\title{HFS: Holistic Query-Aware Frame Selection for Efficient Video Understanding}

\author{Yiqing Yang}
\orcid{0009-0007-2455-4584}
\affiliation{%
  \institution{University of Virginia}
  \city{Charlottesville}
  \state{Virginia}
  \country{USA}}
\email{yiqing.yang@virginia.edu}

\author{Yun Li}
\orcid{0009-0000-5138-1374}
\affiliation{%
  \institution{Hong Kong Polytechnic University}
  \country{Hong Kong}}
\email{yun-eie.li@connect.polyu.hk} 

\author{Daiqing Qi}
\orcid{0000-0001-9543-5792}
\affiliation{%
  \institution{University of Virginia}
  \city{Charlottesville}
  \state{Virginia}
  \country{USA}}
\email{daiqing.qi@virginia.edu}

\author{Lehan Yang}
\orcid{0000-0003-1617-6396}
\affiliation{%
  \institution{University of Virginia}
  \city{Charlottesville}
  \state{Virginia}
  \country{USA}}
\email{lehan.yang@virginia.edu}

\author{Tianlong Wang}
\orcid{0009-0002-0498-6682}
\affiliation{%
  \institution{University of Virginia}
  \city{Charlottesville}
  \state{Virginia}
  \country{USA}}
\email{tianlongwang@virginia.edu}

\author{Wenhao Zhang}
\orcid{0009-0007-2866-1969}
\affiliation{%
  \institution{University of Virginia}
  \city{Charlottesville}
  \state{Virginia}
  \country{USA}}
\email{wenhaozhang@virginia.edu}

\author{Sheng Li}
\orcid{0000-0003-1205-8632}
\affiliation{%
  \institution{University of Virginia}
  \city{Charlottesville}
  \state{Virginia}
  \country{USA}}
\email{shengli@virginia.edu}

\author{Kin-man Lam}
\orcid{0000-0002-0422-8454}
\affiliation{%
  \institution{Hong Kong Polytechnic University}
  \country{Hong Kong}}
\email{enkmlam@polyu.edu.hk}

\renewcommand{\shortauthors}{Yiqing Yang et al.}

\begin{abstract}
  Key frame selection is essentially a set-level optimization problem: the quality of the selected subset depends on the interactions among frames, rather than the score of any single frame. Existing methods generally exhibit three major limitations. Point-wise methods score each frame independently and ignore inter-frame dependencies. Although the training-free set-level methods explicitly model the inter-frame relationships, their selection criteria are fixed and cannot be adapted through downstream task feedback. Learnable methods can leverage data-driven training; however, they lack an explicit, differentiable set-quality objective and rely on offline-generated supervision signals. To address these limitations, we propose an end-to-end trainable and task-adaptive framework for frame selection. A Chain-of-Thought prompt conditions a Small Language Model (SLM) to extract task-specific latent query vectors, which are combined with multimodal features to enable dynamic, query-aware frame scoring. We further formulate a continuous set-level objective function that jointly accounts for relevance, coverage, and redundancy, enabling differentiable set-level optimization via Gumbel-TopK for selecting optimal frame combinations. Finally, we employ a student-teacher mutual learning strategy, in which the student selector (SLM) and teacher reasoner (MLLM) are trained to align their frame-importance distributions via KL divergence. Combined with cross-entropy loss, this design enables fully end-to-end optimization, eliminating reliance on static pseudo-labels. Experiments across multiple benchmarks, including Video-MME, LongVideoBench, MLVU, and NExT-QA, demonstrate that our method significantly outperforms existing frame-selection approaches.
\end{abstract}



\maketitle

\section{Introduction}
\label{sec:intro}
\begin{figure}[t]
  \includegraphics[width=\columnwidth]{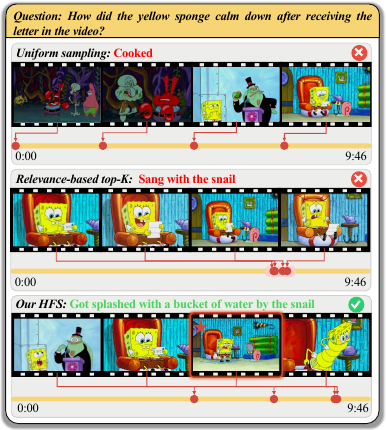}
  \Description{Three sets of sampled video frames labeled (a) uniform sampling, (b) top-K selection, and (c) HFS selection, where only HFS captures the key moment of a character being splashed with water.}
  \caption{Comparison on an event reasoning task. The task requires the model to locate a sparse, critical moment within a nine-minute video. (a) Uniform sampling sampled frames that completely missed the critical information, leading to a wrong prediction. (b) Relevance-based top-K selection identifies frames that are highly relevant to the query. However, it selects highly redundant frames while overlooking the key event necessary to answer the question. (c) Through CoT-conditioned query understanding and set-level optimization, our HFS approach effectively suppresses redundancy and accurately locates the action of being splashed with water, guiding the model to produce the correct answer.}
  \label{fig:ab}
\end{figure}
Multimodal Large Language Models (MLLMs) have made significant progress across modalities~\cite{liu2023llava, 10.5555/3618408.3619222, NEURIPS2022_960a172b, yang2025emoq}, including image understanding, video analysis, and cross-modal reasoning. Unlike static images, video data contains dense temporal information and frames with a high degree of redundancy, which can overwhelm the model's context window and hinder effective processing. This mismatch leads to context overflow when processing lengthy videos, with even moderately sized clips often exceeding the model's processing capabilities. Consequently, selecting informative key frame subsets from raw video has become an essential step in video understanding tasks. 

Researchers have proposed various frame selection strategies to address this challenge. One class employs heuristic sampling methods~\cite{lin-etal-2024-video, zhang2025llavavideo}, such as uniform sampling at fixed intervals. Another approach introduces learnable scoring mechanisms, assigning importance scores to each frame before applying top-K selection to obtain a critical frame subset~\cite{Hu_2025_CVPR, 11092902}. Beyond per-frame scoring, Frame-Voyager~\cite{DBLP:conf/iclr/YuJWCJZXSZWZS25} formulates frame selection as a combinatorial ranking problem, and FFS~\cite{11093988} introduces a flexible selection operation that learns to determine how many frames to retain for each query. Further work leverages MLLM to generate spatial-temporal pseudo-labels offline~\cite{Hu_2025_CVPR}, which are then used to supervise the training of lightweight selectors. Although these methods are effective in specific scenarios, fundamental limitations remain.

Frame selection is inherently a set-level problem, since the quality of the selected subset depends on the interactions among frames rather than individual relevance alone. Existing methods fall short of this view. Point-wise selectors score frames independently~\cite{11092902, Zhang_2025_ICCV}, while temporal-coverage heuristics or random sampling~\cite{lin-etal-2024-video, zhang2025llavavideo} reduce redundancy without modeling semantic interactions. The recent training-free set-level method~\cite{Sun_2025_ICCV} does capture inter-frame diversity, but through a frozen kernel that cannot adapt to downstream feedback, while other learnable selectors~\cite{Hu_2025_CVPR, DBLP:conf/iclr/YuJWCJZXSZWZS25} rely on offline supervision disconnected from the task goal. Figure~\ref{fig:ab} shows a typical failure case: a relevance-based method selects redundant frames of the same event while omitting the decisive moment required to answer the question.

In order to bridge these gaps, we propose HFS, a Holistic, query-aware Frame Selection framework. To the best of our knowledge, HFS is the first to integrate a differentiable, coverage-based set-quality objective with diminishing returns into the end-to-end training of a frame selector jointly with its downstream reasoner. Unlike prior differentiable selectors, which combine per-frame relevance with a pairwise distinctiveness penalty, and unlike training-free set-level selectors, which are neither differentiable nor trained jointly with the reasoner, HFS couples a coverage-driven set objective with online mutual learning. HFS contains a lightweight Small Language Model (SLM) acting as a student frame selector, and an MLLM serving as a teacher video reasoner. Inspired by set-function optimization~\cite{pmlr-v37-wei15,ijcai2018p379,7298928}, the set-quality objective serves as a regularizer during training, enabling the model to learn what constitutes an effective frame subset directly from the downstream task feedback. To provide accurate scoring inputs for set-level optimization, we design a Chain-of-Thought (CoT)-conditioned latent query extraction mechanism~\cite{NEURIPS2022_9d560961}: the SLM is conditioned on a CoT prompt to capture task intent, yields a set of task-specific latent query vectors, and integrates them with multimodal features through a LoRA adapter~\cite{hu2022lora}. To achieve fully end-to-end training without reliance on static offline supervision, we adopt a mutual learning mechanism~\cite{Zhang_2018_CVPR}, where the student selector and teacher reasoner align the frame-importance distributions through KL divergence, jointly optimized with the cross-entropy loss of downstream tasks. This unified objective eliminates reliance on fixed pseudo-labels. Together, these three components, task-aware query understanding, differentiable set-level optimization, and student-teacher mutual learning, form a unified framework in which frame selection and downstream reasoning are jointly optimized, effectively closing the long-standing gap between frame selection quality and task performance.

The main contributions of this paper are as follows:

1. We propose a frame selection framework that integrates a differentiable, coverage-based set-quality objective with diminishing returns into end-to-end training, jointly optimizing the frame selector with the downstream reasoner. To our knowledge, this specific combination has not been adopted by prior differentiable or training-free set-level selectors.

2. We design a CoT-conditioned latent query extraction mechanism that produces task-specific latent query vectors for deep task-intent understanding, along with a mutual learning paradigm that eliminates dependence on static offline supervision.

3. Experiments on Video-MME, MLVU, LongVideoBench, and NExT-QA show that our method significantly outperforms existing approaches, specifically, improving MLVU M-AVG by up to 4.0 percentage points and Video-MME overall accuracy by 2.6 percentage points over the baseline, while surpassing the strongest training-free method by up to 2.0 points.

\section{Related Work}
\label{sec:related}

\noindent\textbf{Video Understanding Using MLLM.}
Early MLLMs extended image understanding capabilities to video inputs through visual alignment~\cite{lin-etal-2024-video}. Share\-GPT4\-Video~\cite{chen2025sharegpt4video} and LLaVA-Video~\cite{zhang2025llavavideo} synthesize dense video descriptions, while MVBench~\cite{li2024mvbench} provides a comprehensive evaluation benchmark, and InternVideo2~\cite{10.1007/978-3-031-73013-9_23} further scales unified multimodal video representations. Long-form video understanding has been addressed using sparse memory mechanisms~\cite{10657734}, temporal modeling~\cite{10656135}, context length extension~\cite{DBLP:conf/iclr/ChenXLHZLFTYLHY25}, and adaptive compression~\cite{shen2025longvu}. Recent studies also explore fine-grained spatial understanding~\cite{11092641, yuan2025videorefer}, enhancing regional-level comprehension in videos.

\noindent\textbf{Video Frame Selection Using MLLM.}
Processing all frames in long videos incurs high computational costs~\cite{Hu_2025_CVPR}, while uniform sampling often discards critical information. Consequently, recent methods have shifted toward adaptive frame selection. AKS~\cite{11092902} balances relevance and coverage, M-LLM~\cite{Hu_2025_CVPR} introduces spatial-temporal pseudo-labels, and FFS~\cite{11093988} learns how many frames to retain. VideoTree~\cite{11094563} uses hierarchical clustering, while Frame-Voyager~\cite{DBLP:conf/iclr/YuJWCJZXSZWZS25} and Q-Frame~\cite{Zhang_2025_ICCV} explore query-aware selection. \text{\small M}DP$^3$~\cite{Sun_2025_ICCV} uses determinantal point processes to perform 
training-free list-wise frame selection. None of these methods, however, optimizes a differentiable set-quality objective jointly with the reasoner, whether because it scores frames independently, fixes the set-level criterion through a frozen kernel, or relies on offline supervision detached from the task. Our online, end-to-end framework closes this gap through mutual learning.

\section{Holistic Query-Aware Frame Selection}

\subsection{Overview}
\begin{figure*}[t]
  \centering
  \includegraphics[width=\textwidth]{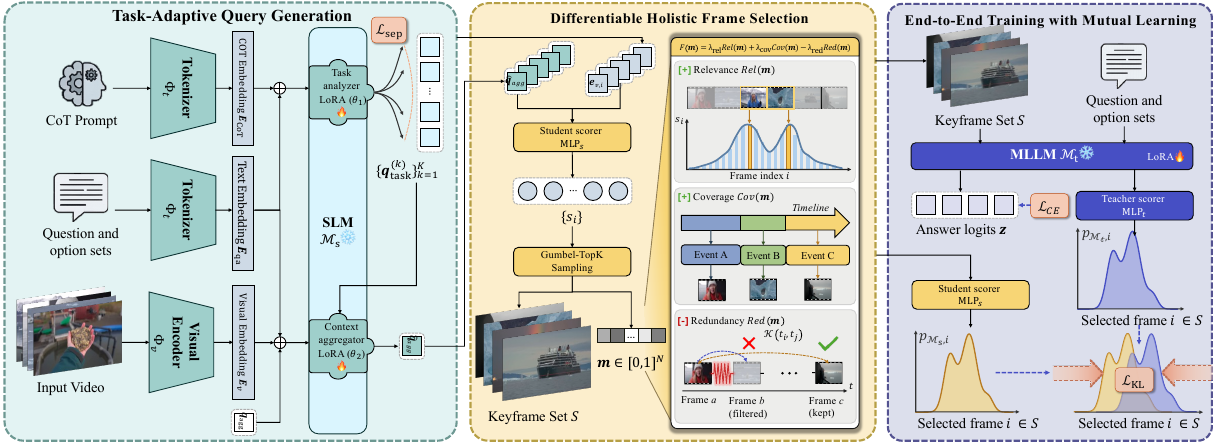}
  \Description{A diagram showing the HFS framework with three stages: task-adaptive query generation using a student model conditioned on a Chain-of-Thought prompt, differentiable holistic frame selection with Gumbel-TopK sampling, and teacher reasoning with KL divergence distillation.}
  \caption{Overall architecture of the proposed HFS framework. 1) In the task-adaptive query generation stage, the student model is conditioned on a CoT prompt to extract task query vectors, which are aggregated with video and text features to form a context-aware query vector. 2) Differentiable holistic frame selection scores each frame and selects key frames via Gumbel-TopK sampling, using a set-level objective as regularization. 3) In the teacher-reasoning stage, the selected frames are fed into the teacher model, with the teacher and student aligning their frame-importance distributions through KL divergence.}
  \label{fig:details}
\end{figure*}
The overall architecture of our holistic frame selection (HFS) framework is illustrated in Figure~\ref{fig:details}. The design principle of HFS is to formulate frame selection as a set-level optimization problem that can be jointly trained end-to-end with downstream tasks. This requires three components: (i) CoT-conditioned query understanding, which goes beyond shallow feature matching to capture task intent; (ii) a differentiable set-quality objective that evaluates the selected subset holistically; and (iii) online training signals from downstream tasks to guide and refine the frame selection process. Let the input video be $\mathcal{V} = \{\mathcal{V}_i\}_{i=1}^N$, where $\mathcal{V}_i$ is the $i$-th frame and $N$ is the total number of frames. We obtain the video embedding matrix $\bm{E}_v$ using a pre-trained visual encoder $\Phi_v$, followed by a linear projection layer $\bm{W}_v$, as follows:
\begin{equation}
\bm{E}_v = [\bm{e}_{v,1}, \dots, \bm{e}_{v,N}]^\top \in \mathbb{R}^{N \times d},
\end{equation}
where $\bm{e}_{v,i} = \bm{W}_v \Phi_v(\mathcal{V}_i) \in \mathbb{R}^d$ is the $d$-dimensional embedding vector of the $i$-th frame. For the text input, let $Q$ denote the question and $\{O_j\}$ denote the set of answer options. We use the language model's embedding layer $\Phi_{t}$ to transform the tokenized text into the text embedding matrix $\bm{E}_{\text{qa}}$, as follows:
\begin{equation}
\bm{E}_\text{qa} = \Phi_{t}([Q; \{O_j\}]) \in \mathbb{R}^{L \times d},
\end{equation}
where $L$ is the length of the tokenized text sequence.

The student frame selector $\mathcal{M}_s$ is an SLM that takes the full video frame embedding matrix $\bm{E}_v$ and the question embedding $\bm{E}_{\text{qa}}$ as input. It analyzes the global visual-textual context and, through a lightweight student scorer $\text{MLP}_s$, predicts an importance score for each frame in the video, thereby identifying a highly representative key frame subset $S$.

This subset is then fed, together with the original question, into the teacher video reasoner $\mathcal{M}_t$. The teacher reasoner is an MLLM that performs the final reasoning task based solely on this compact and informative input. To enable mutual learning, $\mathcal{M}_t$ also maintains a lightweight teacher scorer $\text{MLP}_t$ that produces its own frame-importance distribution from its internal representations, enabling bidirectional alignment between teacher and student during training.

\begin{table*}[t]
\centering
\caption{Performance comparison of video understanding models across Video-MME, the test set of MLVU, and the validation set of LongVideoBench. $^\dagger$ indicates results evaluated by us under the same experimental setting. \colorbox{lightpurple}{Highlighted rows} are our method.}
\label{tab:model_comparison}
\resizebox{\textwidth}{!}{
\begin{tabular}{l c c c c c c c c}
\toprule
\multirow{2}{*}{\textbf{Model}} & \multirow{2}{*}{\textbf{LLM size}} & \multirow{2}{*}{\textbf{\# Frames}} & \multicolumn{4}{c}{\textbf{Video-MME (w.o./w. sub.)}} & \multirow{2}{*}{\textbf{LongVideoBench}} & \multicolumn{1}{c}{\textbf{MLVU}} \\
\cmidrule(lr){4-7}\cmidrule(lr){9-9}
& & & Short & Medium & Long & Overall & & M-AVG \\
\rowcolor{gray!20}
\textit{Avg. Duration} & \textit{} & \textit{} & \textit{1.3 min} & \textit{9 min} & \textit{41 min} & \textit{17 min} & \textit{12 min} & \textit{12 min} \\
\midrule
ShareGPT4Video~\cite{chen2025sharegpt4video} & 8B & 16 & 48.3/53.6 & 36.3/39.3 & 35.0/37.9 & 39.9/43.6 & 39.7 & 33.8 \\
VideoLLaMA2~\cite{cheng2024videollama2} & 7B & 8 & - & - & - & 45.1/46.6 & - & 45.6 \\
Video-XL~\cite{shu2025videoxl} & 7B & 128 & 64.0/67.4 & 53.2/60.7 & 49.2/54.9 & 55.5/61.0 & 50.7 & 45.5 \\
VideoChat2-Mistral~\cite{li2024mvbench} & 7B & 16 & 48.3/52.8 & 37.0/39.4 & 33.2/39.2 & 39.5/43.8 & 39.3 & - \\
Kangaroo~\cite{liu2024kangaroo} & 8B & 64 & 66.1/68.0 & 55.3/55.4 & 46.6/49.3 & 56.0/57.6 & 54.2 & - \\
VILA-1.5~\cite{lin2024vila} & 40B & 14 & - & - & - & - & - & 44.2 \\
LongVA~\cite{zhang2024longva} & 7B & 128/256 & 61.1/61.6 & 50.4/53.6 & 46.2/47.6 & 52.6/54.3 & - & 41.1 \\
LLaVA-OneVision~\cite{li2024llavaone} & 7B & 32 & - & - & - & 58.2/61.5 & 56.3 & - \\
Video-LLaVA~\cite{lin-etal-2024-video} & 7B & 8 & 45.3/46.1 & 38.0/40.7 & 36.2/38.1 & 39.9/41.6 & 39.1 & 30.7 \\
Chat-UniVi-v1.5~\cite{jin2024chatunivi} & 7B & 64 & 45.7/51.2 & 40.3/44.6 & 35.8/41.8 & 40.6/45.9 & - & - \\
Video-CCAM~\cite{fei2024videoccam} & 14B & 96 & 62.2/66.0 & 50.6/56.3 & 46.7/49.9 & 53.2/57.4 & - & 42.9 \\
\midrule
Qwen2.5-VL~\cite{bai2025qwen25vl}$^\dagger$ & 7B & 16 & 63.8/68.7 & 50.3/55.2 & 45.1/51.3 & 53.1/58.4 & 54.5 & 41.8 \\
+ TopK+Threshold$^\dagger$ & 7B & 16 & 65.6/69.6 & 55.1/58.9 & 47.9/52.4 & 56.2/60.3 & 56.5 & 43.4 \\
+ AKS~\cite{11092902}$^\dagger$ & 7B & 16 & 65.9/68.7 & 56.1/59.1 & 48.7/53.1 & 56.9/60.3 & 57.2 & 43.6 \\
+ Q-Frame~\cite{Zhang_2025_ICCV}$^\dagger$ & 7B & 16 & 62.8/65.8 & 54.9/58.4 & 49.1/53.0 & 55.6/59.1 & 57.2 & 39.6 \\
+ \text{\small M}DP$^3$~\cite{Sun_2025_ICCV}$^\dagger$ & 7B & 16 & 65.6/69.9 & 56.1/59.1 & 49.9/52.9 & 57.2/60.6 & 57.0 & 44.4 \\
\rowcolor{lightpurple}
+ \textbf{HFS} & 7B+1.5B & 16 & \textbf{68.9/72.2} & \textbf{56.3/60.4} & \textbf{53.9/55.0} & \textbf{59.7/62.6} & \textbf{57.3} & \textbf{45.6} \\
\midrule
InternVL3~\cite{zhu2025internvl3}$^\dagger$ & 8B & 16 & 71.1/75.3 & 58.8/62.1 & 52.2/53.8 & 60.7/63.7 & 56.7 & 46.0 \\
+ TopK+Threshold$^\dagger$ & 8B & 16 & 71.6/75.8 & 59.8/62.6 & 51.2/57.4 & 60.9/65.3 & 57.2 & 47.6 \\
+ AKS~\cite{11092902}$^\dagger$ & 8B & 16 & 70.1/74.0 & 59.9/61.3 & 51.6/56.9 & 60.5/64.1 & 58.5 & 48.0 \\
+ Q-Frame~\cite{Zhang_2025_ICCV}$^\dagger$ & 8B & 16 & 69.0/73.9 & 59.3/61.8 & 51.1/58.0 & 59.8/64.6 & 57.1 & 48.6 \\
+ \text{\small M}DP$^3$~\cite{Sun_2025_ICCV}$^\dagger$ & 8B & 16 & 71.6/76.0 & \textbf{60.9/63.7} & 52.0/57.2 & 61.5/65.6 & 59.3 & 48.0 \\
+ VideoTree~\cite{11094563}$^\dagger$ & 8B & 16 & - & - & - & 62.6/- & 59.3 & - \\
\rowcolor{lightpurple}
+ \textbf{HFS} & 8B+1.5B & 16 & \textbf{73.8/76.8} & 59.8/63.0 & \textbf{56.4/58.7} & \textbf{63.3/66.1} & \textbf{60.2} & \textbf{50.0} \\
\bottomrule
\end{tabular}
}
\end{table*}

\begin{table}[t]
\centering
\caption{Performance comparison on the NExT-QA benchmark. $^\dagger$ indicates results evaluated by us under the same experimental setting.}
\label{tab:nextqa_comparison}
\resizebox{\columnwidth}{!}{
\begin{tabular}{lccr}
\toprule
\textbf{Model} & \textbf{LLM size} & \textbf{\# Frames} & \textbf{NExT-QA} \\
\rowcolor{gray!20}
\textit{Avg. Duration} & \textit{} & \textit{} & \textit{44 sec} \\
\midrule
LVNet~\cite{park2024lvnet} & 7B & 12 & 71.1 \\
SlowFast-LLaVA~\cite{xu2024slowfast} & 7B & 50 & 64.2 \\
LLaVA-NeXT-Video~\cite{zhang2024llavanext} & 7B & 16 & 62.4 \\
LLaVA-OneVision~\cite{li2024llavaone} & 7B & 32 & 79.4 \\
Tarsier~\cite{wang2024tarsier} & 7B & 8 & 71.6 \\
NVILA~\cite{liu2024nvila} & 8B & 256 & 82.2 \\
Video-XL~\cite{shu2025videoxl} & 7B & - & 77.2 \\
Oryx-1.5~\cite{liu2025oryx} & 7B & 64/256 & 81.8 \\
Qwen2-VL~\cite{wang2024qwen2vl} & 7B & - & 77.6 \\
Qwen2-VL~\cite{wang2024qwen2vl} + M-LLM~\cite{Hu_2025_CVPR} & 7B+1.5B & - & 78.4 \\
\midrule
Qwen2.5-VL~\cite{bai2025qwen25vl}$^\dagger$ & 7B & 16 &77.8 \\
+ TopK+Threshold$^\dagger$ & 7B & 16 & 78.6 \\
+ AKS~\cite{11092902}$^\dagger$ & 7B & 16 & 78.8 \\
+ Q-Frame~\cite{Zhang_2025_ICCV}$^\dagger$ & 7B & 16 & 79.0 \\
+ \text{\small M}DP$^3$~\cite{Sun_2025_ICCV}$^\dagger$ & 7B & 16 & 79.1 \\
\rowcolor{lightpurple}
+ \textbf{HFS} & 7B+1.5B & 16 & \textbf{79.4} \\
\midrule
InternVL3~\cite{zhu2025internvl3}$^\dagger$ & 8B & 16 & 81.8 \\
+ TopK+Threshold$^\dagger$ & 8B & 16 & 80.7 \\
+ AKS~\cite{11092902}$^\dagger$ & 8B & 16 & 80.6 \\
+ Q-Frame~\cite{Zhang_2025_ICCV}$^\dagger$ & 8B & 16 & 82.1 \\
+ \text{\small M}DP$^3$~\cite{Sun_2025_ICCV}$^\dagger$ & 8B & 16 & 81.0 \\
\rowcolor{lightpurple}
+ \textbf{HFS} & 8B+1.5B & 16 & \textbf{83.1} \\
\bottomrule
\end{tabular}
}
\end{table}

\subsection{Adaptive Query Generation}
Some existing video frame selection methods~\cite{Hu_2025_CVPR} typically rely on a static, learnable query vector to aggregate spatio-temporal information. However, this approach is often inadequate for handling diverse video understanding tasks with varying semantic intents. To overcome this limitation, we design a two-stage adaptive query generation mechanism. This mechanism first decouples multidimensional task intentions from the input text, then deeply integrates these intentions with visual context to provide dynamic and precise guidance for subsequent frame selection.

\noindent\textbf{CoT-conditioned latent query extraction.} The objective of this stage is to enable the student selector $\mathcal{M}_s$ to understand the intent behind textual queries. We utilize an SLM with a task analyzer LoRA adapter~\cite{hu2022lora} to specifically process textual information, with parameters denoted as $\theta_1$. To guide the student model $\mathcal{M}_s$ toward deep comprehension of query intent, we define a structured Chain-of-Thought (CoT) prompt~\cite{NEURIPS2022_9d560961}, denoted as $\mathcal{P}_\text{CoT}$. This prompt instructs the model to perform logical analysis based on the question and its candidate options, and to identify key concepts that differentiate these options. As a result, the prompt conditions the model's hidden states to encode multiple task-relevant facets, which we subsequently extract as latent query vectors. The CoT prompt is fed into the text encoder $\Phi_{t}$ to obtain its embedding $\bm{E}_\text{CoT}$, as follows:
\begin{equation}
\bm{E}_\text{CoT} = \Phi_{t}(\mathcal{P}_\text{CoT}) \in \mathbb{R}^{L_p \times d},
\end{equation}
where $L_p$ denotes the token length of $\mathcal{P}_\text{CoT}$. We then concatenate $\bm{E}_{\text{CoT}}$ with the question-answer embedding $\bm{E}_{\text{qa}}$ along the sequence dimension to form the joint input embedding $\bm{E}_{\text{input}}$, as follows:
\begin{equation}
\bm{E}_\text{input} = [\bm{E}_\text{CoT} ; \bm{E}_\text{qa}] \in \mathbb{R}^{(L_p + L) \times d}.
\end{equation}
This combined embedding $\bm{E}_{\text{input}}$ is then fed into the SLM $\mathcal{M}_s$, parameterized by $\theta_1$. Under the guidance of the $\bm{E}_{\text{CoT}}$ component, $\mathcal{M}_s$ produces a final sequence of hidden states $\bm{H} \in \mathbb{R}^{(L_p + L) \times d}$, which encodes the task-relevant facets elicited by the augmented text input. We then sample $K$ vectors from the hidden state sequence $\bm{H}$ at uniformly distributed positions, forming a set of task-specific latent query vectors $\{\bm{q}_\text{task}^{(k)}\}_{k=1}^K$:
\begin{equation}
\begin{split}
\{\bm{q}_\text{task}^{(k)}\}_{k=1}^K &= {\bm{H}[j_k]}_{k=1}^K, \\
\text{where} \quad j_k &= \lfloor \frac{k-1}{\max(K-1,\,1)}(L_p + L - 1) \rfloor .
\end{split}
\end{equation}
Here, $\bm{q}_\text{task}^{(k)} \in \mathbb{R}^d$ denotes the $k$-th task query vector, and its index $j_k$ is determined via linear interpolation over the sequence length $L_p + L$. This uniform sampling strategy serves as a simple yet effective heuristic for capturing information from different stages of the CoT-conditioned hidden state sequence. The resulting set of vectors is intended to capture diverse semantic aspects of the original text query, enabling more expressive and task-aware frame selection.

\noindent\textbf{Query diversification via separation loss.} To ensure these $K$ query vectors capture task requirements from distinct angles rather than converging to similar representations, we introduce a separation loss $\mathcal{L}_{\text{sep}}$. This loss function aims to maximize the angular separation between query vectors by minimizing the sum of squared cosine similarities between pairs:
\begin{equation}
\mathcal{L}_{\text{sep}} = \frac{2}{K(K-1)} \sum_{i=1}^{K} \sum_{j=i+1}^{K} \left( \frac{\bm{q}_\text{task}^{(i)} \cdot \bm{q}_\text{task}^{(j)}}{\|\bm{q}_\text{task}^{(i)}\| \|\bm{q}_\text{task}^{(j)}\|} \right)^2.
\end{equation}

\noindent\textbf{Generation of aggregator query.} After generating text-driven task queries, the next step is to fuse them with global visual information to form an aggregator query. We continue using the same SLM $\mathcal{M}_s$, but perform this multimodal fusion task via the context aggregator LoRA adapter, with parameters denoted as $\theta_2$. Specifically, we construct a concatenated sequence $\bm{E}_{\text{fused}}$ consisting of the video frame embeddings $\bm{E}_v$, text embeddings $\bm{E}_{\text{qa}}$, the $K$ task query vectors $\{\bm{q}_\text{task}^{(k)}\}_{k=1}^K$ generated in the first stage, and a learnable aggregator query token $\bm{q}_\text{agg} \in \mathbb{R}^d$:
\begin{equation}
\bm{E}_\text{fused} = [\bm{E}_v ; \bm{E}_\text{qa} ; \{\bm{q}_\text{task}^{(k)}\}_{k=1}^K ; \bm{q}_\text{agg}] \in \mathbb{R}^{(N + L + K + 1) \times d}.
\end{equation}
In this extended sequence $\bm{E}_\text{fused}$, the aggregator query token $\bm{q}_{\text{agg}}$ is placed at the end to integrate information from all modalities and task queries through the self-attention mechanism of the SLM. The hidden state corresponding to this aggregated token in the final output layer of the SLM is taken as the fused query vector $\hat{\bm{q}}_{\text{agg}}$, as follows:
\begin{equation}
\hat{\bm{q}}_{\text{agg}} = \text{last}(\mathcal{M}_s(\bm{E}_\text{fused}; \theta_2)) \in \mathbb{R}^d.
\end{equation}

To overcome the ``myopic'' nature of traditional top-K selection, which often results in selecting redundant frames, we reformulate frame selection as a set-level optimization problem. Instead of optimizing relevance scores for individual frames, our objective is to directly optimize the quality of the selected set $S$. Drawing inspiration from set-function optimization~\cite{pmlr-v37-wei15,ijcai2018p379,7298928}, which embodies the principle of diminishing returns, we design a differentiable, holistic objective function $F(\bm{m})$. This principle implies that once an event is adequately covered by frames in $S$, the marginal benefit of adding more highly similar frames diminishes sharply.  Accordingly, our objective balances three complementary factors: maximizing task relevance, ensuring information coverage, and minimizing temporal redundancy.

\noindent\textbf{Context-aware relevance scoring.}
We aim to generate a relevance score set $\{s_i\}_{i=1}^N$ for each frame, which serves as the basis for subsequent differentiable sampling. An accurate score must be context-dependent. To this end, we utilize the previously generated aggregator query vector $\hat{\bm{q}}_{\text{agg}}$, which encodes rich information about specific task intent and multimodal context. We concatenate $\hat{\bm{q}}_{\text{agg}}$ with each frame embedding $\bm{e}_{v,i} \in \mathbb{R}^d$ and feed the result into a lightweight student MLP scorer $\text{MLP}_s$. This scorer outputs an initial relevance score $s_i$ for each frame:
\begin{equation}
s_i = \sigma(\text{MLP}_s([\bm{e}_{v,i} ; \hat{\bm{q}}_{\text{agg}}])) \in [0, 1], \quad \forall i \in \{1, \dots, N\},
\end{equation}
where $\sigma$ denotes the sigmoid function, used to constrain the score to the interval $[0, 1]$. The discrete operation of selecting the top-K frames from the score set $\{s_i\}$ is non-differentiable. To relax this process, we employ the Gumbel-TopK technique~\cite{JMLR:v21:19-985}, which introduces Gumbel noise and generates a continuous, differentiable selection mask $\bm{m}=\left\{m_i\right\}_{i=1}^N\  \in [0,1]^N$ via the Softmax function, where $m_i$ represents the ``soft'' probability of selecting the $i$-th frame:
\begin{equation}
\bm{m}, S = \text{Gumbel-TopK}(\{s_i\}_{i=1}^N, \tau, k_\text{sel}),
\end{equation}
where $k_{\text{sel}}$ is the number of frames to select, $S$ is the index set of selected frames, and $\tau$ is the temperature parameter, which is gradually annealed during training.

\noindent\textbf{Holistic set objective.}
We define a continuous objective function $F(\bm{m})$ to evaluate the quality of the soft-selected set based on three complementary criteria: Relevance (Rel), Coverage (Cov), and Redundancy (Red).

$\text{Rel}(\bm{m})$ measures the expected cumulative relevance score of the selected frames. It is computed by weighting the original relevance scores $s_i$ with the soft mask $\bm{m}$:
\begin{equation}
\text{Rel}(\bm{m}) = \sum_{i=1}^{N} s_i \cdot m_i,
\end{equation}

$\text{Cov}(\bm{m})$ encourages at least one selected frame to achieve a high relevance score, preventing the selection from distributing probability mass uniformly across low-relevance frames. We implement this term using the log-sum-exp function, which serves as a smooth approximation of the maximum, with temperature parameter $\tau_c$:
\begin{equation}
\text{Cov}(\bm{m}) = \tau_c \log \left( \sum_{i=1}^{N} \exp\left(\frac{s_i \cdot m_i}{\tau_c}\right) \right).
\end{equation}

To promote diversity, we introduce a temporal redundancy term $\text{Red}(\bm{m})$, which is minimized during optimization. We define a temporal similarity kernel function $\mathcal{K}(t_i, t_j)$ that assigns higher redundancy to temporally proximate frames:
\begin{equation}
\mathcal{K}(t_i, t_j) = \exp\left(-\frac{(t_i - t_j)^2}{2\gamma^2}\right),
\end{equation}
\begin{equation}
\text{Red}(\bm{m})= \sum_{i=1}^{N}\sum_{j=1, j \neq i}^{N} m_i \cdot m_j \cdot \mathcal{K}(t_i, t_j).
\end{equation}
Here $t_i$ denotes the normalized index of the $i$-th candidate on the grid of the $N$ uniformly sampled frames, rather than its absolute timestamp. The bandwidth $\gamma$ therefore lives on a scale that is independent of video duration, which is what allows a single $\gamma$ to serve short and long videos alike.

We combine these three components with weights to obtain the continuous set-level function $F(\bm{m})$, as follows:
\begin{equation}
F(\bm{m}) = \lambda_\text{rel} \cdot \text{Rel}(\bm{m}) + \lambda_\text{cov} \cdot \text{Cov}(\bm{m}) - \lambda_\text{red} \cdot \text{Red}(\bm{m}),
\end{equation}
where $\lambda_\text{rel}$, $\lambda_\text{cov}$, and $\lambda_\text{red}$ are hyperparameters that balance the contributions of relevance, coverage, and redundancy, respectively.

Notably, $F(\bm{m})$ acts as a differentiable regularizer during end-to-end training, rather than an independent objective requiring formal approximation guarantees at inference time. Unlike the predefined set-quality criteria used in the training-free method, the scores $\{s_i\}$ in $F(\bm{m})$ are generated by a learnable scorer, which receives downstream gradient signals. This allows the notions of ``relevance'' and ``redundancy'' to adapt dynamically to specific tasks. The total loss function $\mathcal{L}_\text{total}$ includes the term $-\lambda_\text{set} \cdot F(\bm{m})$. Minimizing $\mathcal{L}_\text{total}$ therefore encourages maximization of $F(\bm{m})$, leading to the selection of more informative and diverse frame subsets.
\subsection{End-to-End Training with Mutual Learning}
\label{sec:mutual}
Training a frame selector on pseudo-labels that an MLLM generates offline is suboptimal, because such supervision cannot adapt to task-specific feedback. To address this limitation, we propose an end-to-end online distillation framework~\cite{Zhang_2018_CVPR} in which both the student scorer $\text{MLP}_s$ and the teacher scorer $\text{MLP}_t$ are jointly optimized through a mutual alignment objective $\mathcal{L}_{\text{KL}}$.

\noindent\textbf{Downstream task supervision.} One objective of this framework is to maximize the accuracy of downstream tasks. In our teacher-student architecture, the MLLM $\mathcal{M}_t$ serves as the teacher. It receives the image data $\mathcal{V}_S = \{\mathcal{V}_i\}_{i \in S}$ corresponding to the key frame indices $S$ selected by the student model $\mathcal{M}_s$, along with the original question text $Q$ and the list of options $\{O_j\}$. The teacher $\mathcal{M}_t$ performs inference based on this sparse input and generates answer logits $\bm{z} \in \mathbb{R}^C$, where $C$ is the number of candidate answers:
\begin{equation}
\bm{z} = \mathcal{M}_t(\mathcal{V}_S, Q, \{O_j\}).
\end{equation}
We compute the standard cross-entropy loss $\mathcal{L}_{\text{CE}}$ as one of the objectives:
\begin{equation}
\mathcal{L}_{\text{CE}} = -\sum_{c=1}^{C} y_{c} \log(\text{Softmax}(\bm{z})_c),
\end{equation}
where $y_c$ is the ground-truth label. It forms a one-hot vector $\bm{y} \in \{0,1\}^C$, such that $y_c = 1$ when $c$ is the index of the correct answer, and $y_c = 0$ otherwise. Simultaneously, we introduce the set-level objective $F(\bm{m})$ as a structured regularization term, which directly evaluates the intrinsic quality of the selected set $S$.

\noindent\textbf{Mutual learning.} To enable co-training, we extract frame-level representations from the second-to-last hidden layer of $\mathcal{M}_t$. Vision tokens corresponding to each frame are identified and average-pooled to produce vectors $\bm{h}_i$, forming $\bm{H}_{\mathcal{M}_t} = [\bm{h}_1, \dots, \bm{h}_{k_{\text{sel}}}]^\top \in \mathbb{R}^{k_{\text{sel}} \times d_h}$,
where $d_h$ denotes the hidden dimension of the teacher model $\mathcal{M}_t$. We also extract the final text token's hidden state $\bm{h}_{\text{con}}$ as a global context summary. The teacher scorer $\text{MLP}_t$ then generates its importance distribution:
\begin{equation}
\bm{p}_{\mathcal{M}_t} = \text{Softmax}(\text{MLP}_t([\bm{H}_{\mathcal{M}_t} ; \bm{h}_{\text{con}}])) \in \mathbb{R}^{k_{\text{sel}}}.
\end{equation}

Correspondingly, the student distribution $\bm{p}_{\mathcal{M}_s}$ is obtained by smoothing its own frame importance scores $\{s_i\}_{i \in S}$ using a distillation temperature $\tau_d$:
\begin{equation}
\bm{p}_{\mathcal{M}_s} = \text{Softmax}(\{s_i/\tau_d\}_{i \in S} ) \in \mathbb{R}^{k_\text{sel}}.
\end{equation}

We employ the Kullback-Leibler divergence as our alignment loss $\mathcal{L}_{\text{KL}}$. Gradients from $\mathcal{L}_{\text{KL}}$ flow back to both the student scorer $\text{MLP}_s$ and the teacher scorer $\text{MLP}_t$, forcing them to co-evolve:
\begin{equation}
\mathcal{L}_{\text{KL}} = \sum_{i=1}^{k_{\text{sel}}} \bm{p}_{\mathcal{M}_t, i} \log \frac{\bm{p}_{\mathcal{M}_t, i}}{\bm{p}_{\mathcal{M}_s, i}}.
\end{equation}

\noindent\textbf{Overall objective.} Our final training objective $\mathcal{L}_\text{total}$ is a multi-task loss function that integrates all the above loss components:
\begin{equation}
\mathcal{L}_\text{total} = \mathcal{L}_{\text{CE}} + \lambda_{\text{KL}} \cdot \mathcal{L}_{\text{KL}} - \lambda_{\text{set}} \cdot F(\bm{m}) + \lambda_{\text{sep}} \cdot \mathcal{L}_{\text{sep}},
\end{equation}
where $\lambda_{\text{KL}}$, $\lambda_{\text{set}}$, and $\lambda_{\text{sep}}$ are hyperparameters controlling the contribution of each term.

\section{Experiments}

\subsection{Experimental Setup}
\noindent\textbf{Datasets.}
We train on 250K annotated samples from VideoChat2-IT~\cite{li2024mvbench}, covering description, question answering and reasoning, together with 196K samples from LLaVA-Video-178K~\cite{zhang2025llavavideo}. We evaluate on four benchmarks: Video-MME~\cite{Fu_2025_CVPR}, which is split into short, medium and long subsets and evaluated with and without subtitles; the validation set of LongVideoBench~\cite{Wu_2024_NeurIPS} and the test set of MLVU~\cite{Zhou_2025_CVPR}, both averaging 12 minutes per video; and the test set of NExT-QA~\cite{Xiao_2021_CVPR}, averaging 44 seconds.

\begin{table*}[t]
\centering
\caption{Ablation study on the main components using Qwen2.5-VL-7B-Instruct as the teacher model. \colorbox{lightpurple}{Highlighted rows} indicate the
configuration used in our final model.}
\resizebox{\textwidth}{!}{
\begin{tabular}{lccccccc}
\toprule
\textbf{Selection Method} & \textbf{CoT-Query} & \textbf{Set Objective} & \textbf{KL-Distill} & \textbf{$\mathcal{L}_{\text{sep}} $} & \textbf{MLVU} & \textbf{LongVideoBench} & \textbf{NExT-QA} \\
\midrule
Baseline& $\times$ & $\times$ & $\times$ & $\times$  & 42.2 & 55.0 & 78.1 \\
HFS w/o CoT-Query & $\times$ & $\checkmark$ & $\checkmark$ & $\times$ & 43.4 & 55.3 & 78.5\\
HFS w/o Set Objective & $\checkmark$ & $\times$ & $\checkmark$ & $\checkmark$ & 43.8 & 55.2 & 78.3\\
HFS w/o KL Distillation & $\checkmark$ & $\checkmark$ & $\times$ & $\checkmark$ & 44.4 & 55.9 & 78.7\\
HFS w/o $\mathcal{L}_{\text{sep}} $ & $\checkmark$ & $\checkmark$ & $\checkmark$ & $\times$ & 44.8 & 56.1 & 78.9\\
\rowcolor{lightpurple}
HFS  & $\checkmark$ & $\checkmark$ & $\checkmark$ & $\checkmark$ & \textbf{45.6} & \textbf{57.3} & \textbf{79.4} \\
\bottomrule
\end{tabular}
}
\label{tab:my-ablation}
\end{table*}

\setcounter{topnumber}{1}
\begin{table}[t]
\centering
\caption{Ablation study on the number of generated queries $K$ using InternVL3-8B as the teacher model.}
\label{tab:query-ablation}
\resizebox{\columnwidth}{!}{
\begin{tabular}{l c c c}
\toprule
\textbf{Task analyzer LoRA} & \textbf{Context aggregator LoRA} & \textbf{$K$} & \textbf{Video-MME (w/ sub.)} \\
\midrule
$\times$ & $\checkmark$ & 0 & 64.5 \\
$\mathbf{e}_{\text{qa}} \to \mathbf{q}_{\text{task}}^{(1)}$ & $\checkmark$ & 1 & 65.2 \\
$\mathbf{e}_{\text{qa}} \to \{\mathbf{q}_{\text{task}}^{(k)}\}_{k=1}^{2}$ & $\checkmark$ & 2 & 65.7 \\
\rowcolor{lightpurple}
$\mathbf{e}_{\text{qa}} \to \{\mathbf{q}_{\text{task}}^{(k)}\}_{k=1}^{3}$ & $\checkmark$ & 3 & \textbf{66.1} \\
$\mathbf{e}_{\text{qa}} \to \{\mathbf{q}_{\text{task}}^{(k)}\}_{k=1}^{4}$ & $\checkmark$ & 4 & 65.9 \\
\bottomrule
\end{tabular}
}
\end{table}

\begin{table}[t]
\centering
\caption{Ablation study on the components of the set quality objective $F(\bm{m})$ using InternVL3-8B as the teacher model.}
\label{tab:submodular-ablation}
\resizebox{\columnwidth}{!}{
\begin{tabular}{ccc cc}
\toprule
\multicolumn{3}{c}{\textbf{Set quality Objective}}  & \multirow{2}{*}{\textbf{MLVU}} & \multirow{2}{*}{\textbf{LongVideoBench}}\\
\cmidrule(lr){1-3}
\textbf{Relevance} & \textbf{Coverage} & \textbf{Redundancy} & & \\
\midrule
$\times$ & $\times$ & $\times$ & 47.8 & 58.0\\
$\checkmark$ & $\times$ & $\times$ & 48.4 & 58.5\\
$\checkmark$ & $\checkmark$ & $\times$ & 49.4 & 59.3\\
$\checkmark$ & $\times$ & $\checkmark$ & 49.0 & 59.1\\
\rowcolor{lightpurple}
$\checkmark$ & $\checkmark$ &$\checkmark$& \textbf{50.0} & \textbf{60.2}\\
\bottomrule
\end{tabular}
}
\end{table}

\noindent\textbf{Baselines.}
We compare HFS with frame selection methods evaluated under identical settings. (1) \textbf{Uniform sampling} takes $k_{\text{sel}}$ frames at fixed intervals. (2) \textbf{TopK+Threshold} computes the CLIP cosine similarity between each frame and the query, discards frames below a threshold of $0.85$, and keeps the top $k_{\text{sel}}$ of the remainder. (3) \textbf{AKS}~\cite{11092902}, (4) \textbf{Q-Frame}~\cite{Zhang_2025_ICCV} and (5) \textbf{\text{\small M}DP$^3$}~\cite{Sun_2025_ICCV} are training-free and applied directly at inference time.

\noindent\textbf{Implementation details.}
We employ Qwen2.5-1.5B-Instruct~\cite{qwen2.5} as the student frame selector $\mathcal{M}_s$, and Qwen2.5-VL-7B-Instruct~\cite{bai2025qwen25vl} together with InternVL3-8B-hf~\cite{zhu2025internvl3} as the teacher video reasoner $\mathcal{M}_t$. For video encoding, we use CLIP ViT-Base-Patch32~\cite{Radford_2021_ICML} to extract visual features from input frames, keeping the visual encoder frozen throughout training. The linear projection layer $\bm{W}_v$ is jointly trained with the other learnable components. The initial video is uniformly sampled at a resolution of $224 \times 224$ to yield $ N = 128$ frames, from which the selector identifies $k_{\text{sel}} = 16$ key frames.

We perform parameter-efficient fine-tuning using LoRA~\cite{hu2022lora} on three adapters. In the student frame selector $\mathcal{M}_s$, the adapter for CoT-conditioned latent query extraction, parameterized as $\theta_1$ and yielding $K=3$ task-specific latent query vectors, uses rank $r_1=16$, and the adapter for context aggregation, parameterized as $\theta_2$, uses rank $r_2=16$. In the teacher video reasoner $\mathcal{M}_t$, the adapter for answer generation uses rank $r_3=8$. We optimize all trainable parameters with AdamW~\cite{Loshchilov_2019_ICLR}, a learning rate of $1\times10^{-5}$, a weight decay of $0.01$, and an effective batch size of $16$.

The set-level objective balances its three terms with $\lambda_{\text{rel}}=0.5$, $\lambda_{\text{cov}}=0.3$ and $\lambda_{\text{red}}=0.2$. For Gumbel-TopK sampling we initialize the temperature at $\tau=2.0$ and decay it exponentially by $0.999$ per step down to $\tau_{\text{min}}=0.5$. The smoothing parameter $\tau_c$ of $\text{Cov}(\bm{m})$ is $2.0$ and the temporal kernel bandwidth is $\gamma=10.0$. The alignment weight $\lambda_{\text{KL}}$ increases linearly from $0.1$ to $1.0$ over the first epoch as a warm-up, and the distillation temperature $\tau_d$ is $0.5$. The set-objective regularization weight is $\lambda_{\text{set}}=1\times10^{-4}$ and the query separation weight is $\lambda_{\text{sep}}=0.01$.

\subsection{Main Results}
Throughout this section, \colorbox{lightpurple}{highlighted rows} mark HFS in the comparison tables and the configuration we adopt in the ablation tables. Table~\ref{tab:model_comparison} reports HFS on Video-MME~\cite{Fu_2025_CVPR}, MLVU~\cite{Zhou_2025_CVPR} and Long\-Video\-Bench~\cite{Wu_2024_NeurIPS}, with representative video understanding models in the upper section and frame selection methods under identical settings in the lower section. On Video-MME, HFS improves consistently across the short, medium and long subsets, and InternVL3~\cite{zhu2025internvl3} with HFS achieves the best overall score while leading on both the short and the long subsets. On MLVU, HFS raises the M-AVG score of Qwen2.5-VL~\cite{bai2025qwen25vl} and InternVL3 by 3.8 and 4.0 percentage points. On LongVideoBench, InternVL3 with HFS reaches 60.2\%, consistently ahead of all baseline models. We further compare against VideoTree~\cite{11094563}, the strongest publicly available list-wise selector, re-evaluated under the same InternVL3 setting with a matched 16-frame budget; it reaches 62.6 on Video-MME (without subtitles) and 59.3 on LongVideoBench, both below HFS (63.3 and 60.2).

Table~\ref{tab:nextqa_comparison} evaluates NExT-QA~\cite{Xiao_2021_CVPR}, which stresses fine-grained temporal and causal reasoning and therefore the quality of frame selection. InternVL3~\cite{zhu2025internvl3} with HFS reaches 83.1\%, surpassing all state-of-the-art methods, including NVILA~\cite{liu2024nvila} and Oryx-1.5~\cite{liu2025oryx}.

\setcounter{topnumber}{2}
\subsection{Ablation Studies}
\label{sec:ablation}
As shown in Table~\ref{tab:my-ablation}, we validated the effectiveness of the four primary components in HFS. The baseline is defined as a model trained end-to-end using only $\mathcal{L}_{\text{CE}}$, employing static queries and relying on top-k selections from $\text{MLP}_s$. HFS, which integrates all components, consistently outperforms the baseline across all three benchmarks, demonstrating the overall efficacy of the proposed HFS framework.

\noindent\textbf{Impact of CoT-conditioned latent queries.} We further examine the impact of varying the number of query vectors $K$ in Table~\ref{tab:query-ablation}, using InternVL3-8B~\cite{zhu2025internvl3} as the teacher model for this experiment. When $K=0$, the model degenerates to using static queries, yielding the lowest performance. As $K$ increases from 1 to 3, performance on Video-MME steadily improves, peaking at 66.1\% when $K=3$. However, performance slightly declined to 65.9\% at $K=4$, likely due to the introduction of redundant information. Based on these results, we set $K=3$ for all other experiments.

\noindent\textbf{Set-level objective analysis.}
As shown in Table~\ref{tab:submodular-ablation}, the baseline without $F(\bm{m})$ achieves 47.8\% on MLVU~\cite{Zhou_2025_CVPR}. Adding the relevance term $\text{Rel}(\bm{m})$ alone yields a 0.6 percentage point improvement. Incorporating the coverage term $\text{Cov}(\bm{m})$ on top of $\text{Rel}(\bm{m})$ delivers the greatest marginal gain, indicating that incentivizing coverage is crucial for avoiding information bottlenecks. Simultaneously introducing the redundancy term $\text{Red}(\bm{m})$ further improves performance to 50.0\%.\ A sweep over four weight configurations moves Video-MME accuracy by at most 1.4 points, so the objective does not require delicate tuning. Table~\ref{tab:kernel-ablation} further examines how $\text{Red}(\bm{m})$ measures redundancy. Replacing the temporal kernel with a DINOv2 cosine kernel, or averaging the two, is no better than the temporal prior alone, because a purely semantic kernel also suppresses frames that look alike but carry distinct task evidence. Semantic content already reaches the objective through the learned relevance scores $s_i$ in $\text{Rel}(\bm{m})$ and $\text{Cov}(\bm{m})$, so the redundancy term only needs the cheap temporal prior and does not require a second visual encoder.

\noindent\textbf{Supervision strategy comparison.}
Table~\ref{tab:KL} compares three different supervision strategies: 1) training with static pseudo-labels generated offline by using Qwen2.5-VL-7B-Instruct~\cite{bai2025qwen25vl}; 2) end-to-end training using only cross-entropy loss from the downstream task; and 3) our complete end-to-end online distillation framework. Training with $\mathcal{L}_{\text{CE}}$ but without the $\mathcal{L}_{\text{KL}}$ distillation loss already outperforms using static pseudo-label supervision. By introducing the distillation loss $\mathcal{L}_{\text{KL}}$, our approach achieves state-of-the-art performance on both NExT-QA~\cite{Xiao_2021_CVPR} and Video-MME~\cite{Fu_2025_CVPR}.

\begin{table}[t]
\centering
\caption{Ablation on the redundancy kernel using InternVL3-8B as the teacher model. ``Semantic'' replaces the temporal kernel with a DINOv2 cosine kernel $\frac{1}{2}(1+\cos(\bm{d}_i,\bm{d}_j))$, and ``Hybrid'' averages the two kernels.}
\label{tab:kernel-ablation}
\resizebox{\columnwidth}{!}{
\begin{tabular}{lcc}
\toprule
\textbf{Redundancy kernel} & \textbf{MLVU} & \textbf{LongVideoBench} \\
\midrule
w/o Red & 49.4 & 59.3 \\
Semantic (DINOv2 cosine) & 49.6 & 59.7 \\
Hybrid & 49.8 & 59.8 \\
\rowcolor{lightpurple}
Temporal (ours) & \textbf{50.0} & \textbf{60.2} \\
\bottomrule
\end{tabular}
}
\end{table}

\begin{table}[t]
\centering
\caption{Ablation study on the supervision strategy using InternVL3-8B as the teacher model.}
\label{tab:KL}
\resizebox{\columnwidth}{!}{
\begin{tabular}{cccccc}
\toprule
\textbf{Pseudo-label} & \textbf{$\mathcal{L}_\text{CE}$} & \textbf{$\mathcal{L}_\text{KL}$}& \textbf{NExT-QA}  & \textbf{Video-MME (w/ sub.)} \\
\midrule
$\checkmark$ & $\times$ & $\times$ & 82.1 & 64.4 \\
$\times$ & $\checkmark$ & $\times$ & 82.8 & 65.3 \\
\rowcolor{lightpurple}
$\times$ & $\checkmark$ & $\checkmark$ & \textbf{83.1} & \textbf{66.1} \\
\bottomrule
\end{tabular}
}
\end{table}

\begin{figure}[t]
  \includegraphics[width=\columnwidth]{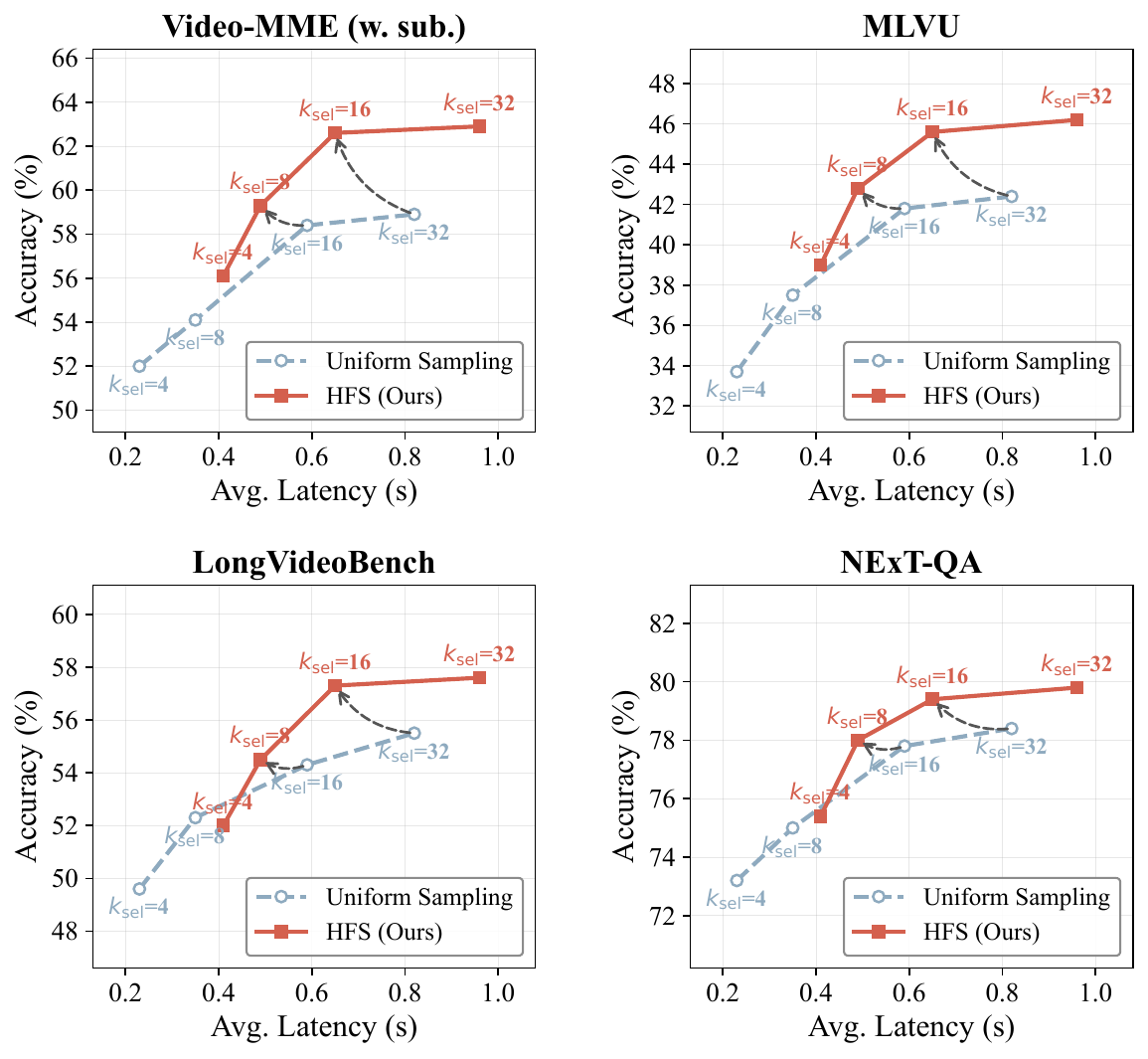}
  \Description{A scatter plot showing accuracy versus latency for uniform sampling and HFS at different numbers of selected frames.}
  \caption{Accuracy-latency comparison of uniform sampling and HFS under different $k_{\text{sel}}$ settings (teacher model: Qwen2.5-VL-7B-Instruct). The dotted arrow indicates that HFS with a smaller $k_{\text{sel}}$ can exceed the accuracy of uniform sampling with a larger $k_{\text{sel}}$, while having lower inference latency.}
  \label{fig:ksel-ablation}
\end{figure}

\noindent\textbf{Efficiency analysis.}
Figure~\ref{fig:ksel-ablation} compares the performance and latency of HFS versus uniform sampling under different frame selection counts $k_{\text{sel}}$. HFS significantly outperforms uniform sampling across all $k_{\text{sel}}$ settings. Notably, HFS with $k_{\text{sel}}=8$ surpasses the accuracy of uniform sampling with $k_{\text{sel}}=16$, while reducing inference latency from 0.59s to 0.49s per sample. Table~\ref{tab:latency} reports per-sample latency for all six selection methods under a single protocol. HFS stays within 0.06s of uniform sampling and runs 2.4 to 3.5 times faster than the other query-aware selectors.{} The selector scores all $N$ candidates in one pass, so its overhead is constant in $k_{\text{sel}}$ and the growth in Figure~\ref{fig:ksel-ablation} comes from the reasoner.

\begin{table}[t]
\centering
\caption{Per-sample inference latency (s) at $k_{\text{sel}}{=}16$, averaged over the four benchmarks, with batch size 1 on a single H100. The figure for HFS includes the student-selector overhead.}
\label{tab:latency}
\resizebox{\columnwidth}{!}{
\begin{tabular}{lcccccc}
\toprule
& \textbf{Uniform} & \textbf{TopK+Thr.} & \textbf{AKS} & \textbf{Q-Frame} & \textbf{\text{\small M}DP$^3$} & \cellcolor{lightpurple}\textbf{HFS} \\
\midrule
\textbf{Latency}$\downarrow$ & 0.59 & 2.25 & 1.61 & 2.29 & 1.53 & \cellcolor{lightpurple}0.65 \\
\bottomrule
\end{tabular}
}
\end{table}

\ifHFSAppendix\else\vskip 0pt plus 1fill\relax\fi
\section{Conclusion}
This paper proposes an end-to-end, task-adaptive framework for video frame selection, where a CoT-conditioned latent query extraction mechanism adapts the selection focus to diverse task intents. The set-level optimization objective jointly enhances relevance and coverage while suppressing redundancy. Moreover, the mutual learning mechanism between the student selector and the teacher reasoner eliminates the reliance on offline pseudo-labels. Extensive experiments across multiple benchmarks demonstrate that our method consistently outperforms existing approaches.
The gains hold across both teacher backbones and all four benchmarks.

\setcounter{topnumber}{4}
\clearpage
\bibliographystyle{ACM-Reference-Format}
\bibliography{main}

\ifHFSAppendix
\clearpage
\raggedbottom
\appendix
\section*{Overview of the Appendix}
The appendix is organized as follows. Appendix~\ref{sec:prompt} gives the exact prompt templates used by the student selector and the teacher reasoner, together with the design rationale behind the CoT prompt. Appendix~\ref{app:impl} covers the answer supervision, the optimization dynamics, the status of the set objective, and the teacher scorer that produces the frame-importance distribution used by the alignment loss. Appendix~\ref{app:additional} collects supporting quantitative analyses that extend the main text, covering training cost, training stability and hyperparameter sensitivity, statistical reliability, query diversity and selection redundancy, robustness to query paraphrasing, and the size of the initial candidate pool. Appendix~\ref{app:benchmark_results} reports per-category results on LongVideoBench and NExT-QA. Appendix~\ref{app:qualitative_analysis} presents qualitative comparisons of the frames selected by each method. Appendix~\ref{app:limitations} closes with the limitations of HFS and the directions they suggest.

\section{Detailed Prompts}
\label{sec:prompt}
\noindent\textbf{CoT prompt for student selector $\mathcal{M}_s$.} Below is the prompt template provided to the student selector $\mathcal{M}_s$ (Qwen2.5-1.5B-In\-struct)~\cite{qwen2.5} to condition it on a Chain-of-Thought (CoT) prompt. A single fixed template is applied automatically by the student selector to each question and option pair, with no human annotation at any stage. Conditioned on it, $\mathcal{M}_s$ produces hidden states from which we sample $K$ latent query vectors. The placeholders \texttt{[Question]} and \texttt{[Options]} are replaced at runtime with specific inputs for each sample, and the same two placeholders carry over to the teacher prompt in Box 2, so a single pair of substitutions drives both stages. Nothing else in either template varies across samples, benchmarks or teachers.

\begin{promptbox}{Box 1: CoT Prompt for Student Selector}
You are a question analysis assistant. Your task is to perform a detailed \textcolor{red}{logical analysis} of the given question and options. This analysis will guide the identification of key decision points.

\textbf{Question:} [Question]

\textbf{Options:} [Options]

\textbf{Task:} Generate a step-by-step logical analysis to identify the key information needed to answer the question.

\textbf{Your analysis should:}
\begin{enumerate}
\item Break down the main question into its core semantic components.
\item Analyze the options and pinpoint the specific visual or temporal evidence required to distinguish between them.
\item Consider what to look for in the video, such as: temporal sequences (e.g., what happens first, then next?), causal relationships (e.g., what action causes another?), entity interactions (e.g., who is doing what to whom?).
\item Write down your reasoning process clearly.
\end{enumerate}

\textbf{Begin your analysis:} [LOGICAL ANALYSIS]
\end{promptbox}

\noindent\textbf{Prompt for teacher reasoner $\mathcal{M}_t$.} Input for the teacher reasoner $\mathcal{M}_t$ (Qwen2.5-VL-7B~\cite{bai2025qwen25vl} or InternVL3-8B~\cite{zhu2025internvl3}) is a structured list of multimodal content. This list combines the $k_\text{sel}=16$ selected key frames with text components, including optional timestamps and the main prompt text, as shown below. The placeholders \texttt{[Question]} and \texttt{[Options]} are replaced with the actual question and options, while $t_i$ represents the timestamp of the $i$-th selected frame. This list is then formatted using the MLLM processor's chat template before being passed to the model.

\begin{promptbox}{Box 2: Prompt for Teacher Reasoner}
{[}

\hspace{5mm}\{``type'': ``image'', ``image'': {[}Selected Frame 1{]}\},

\hspace{5mm}\{``type'': ``text'', ``text'': ``{[}Frame 1 at $t_1${]}''\},

\hspace{5mm}\{``type'': ``image'', ``image'': {[}Selected Frame 2{]}\},

\hspace{5mm}\{``type'': ``text'', ``text'': ``{[}Frame 2 at $t_2${]}''\},

\hspace{5mm}...

\hspace{5mm}\{``type'': ``image'', ``image'': {[}Selected Frame 16{]}\},

\hspace{5mm}\{``type'': ``text'', ``text'': ``{[}Frame 16 at $t_{16}${]}''\},

\hspace{5mm}\{``type'': ``text'', ``text'':

\hspace{10mm}``Based on the selected key frames from the video, answer the following question.

\hspace{10mm}Question: {[}Question{]}

\hspace{10mm}Options: {[}Options{]}

\hspace{10mm}Please select the most appropriate option:''

\hspace{5mm}\}

{]}
\end{promptbox}

\noindent\textbf{Design rationale of the CoT prompt.}
The CoT prompt is deliberately kept concise and structured
around three explicit reasoning dimensions: semantic
decomposition of the question, option-level discrimination,
and anticipation of visual or temporal evidence. This design
encourages the student model to produce hidden states that
simultaneously encode multiple aspects of the task intent,
which in turn yields more diverse and informative implicit
query vectors after uniform sampling. We also experimented
with longer instructions and few-shot exemplars during early
development, but found that the current concise version
offers the best trade-off between the informativeness of the resulting
latent queries and the cost of conditioning, for a 1.5B-parameter student model.

The template is the only prompt involved. It is fixed once and applied automatically by the student model to each question and option pair, so no chain of thought is written by hand for any video and no human annotation enters the pipeline at any stage, either during training or at inference. A trained HFS selector can therefore be deployed directly to unseen videos and unseen questions without collecting any new supervision.

\section{Implementation Details}
\label{app:impl}

\subsection{Answer Supervision}
The downstream loss is defined over the option letters rather than over the full
vocabulary. At the final position of the teacher sequence we take the logits of the
token ids that correspond to the letters A through F, and $\mathcal{L}_{\text{CE}}$ is
computed over that restricted set. This keeps the supervision aligned with the
multiple-choice format of the benchmarks and makes the loss insensitive to how the
teacher would phrase a free-form answer, which would otherwise inject variance that has
nothing to do with frame selection.

Benchmarks differ in how many options a question carries, so a batch is first partitioned
by option count and each group is passed through the teacher separately, which keeps the
restricted option set well defined without padding the option list. The effective batch
size of $16$ is realized as a micro-batch of $2$ with gradient accumulation over $8$ steps.

\subsection{Optimization Dynamics}
The alignment term $\mathcal{L}_{\text{KL}}$ is defined only over the frames that were
selected, so it cannot by itself express a preference for a frame that the selector
never proposed. Taken alone this would risk locking in whichever frames happen to
score well early in training. Two mechanisms prevent that. First, the selection is not
fixed across steps: $\bm{m}$ is re-scored and re-sampled at every step under fresh Gumbel
noise, so a frame that loses once is not excluded from later rounds. Second, the set
objective $F(\bm{m})$ is evaluated through the straight-through relaxation, which
propagates gradients to all $N$ candidates rather than to the $k_{\text{sel}}$ selected
ones, so the scores of unselected frames keep being updated. The warm-up on
$\lambda_{\text{KL}}$ described above works in the same direction, since it keeps the
alignment weak while the selector is still exploring.

\subsection{Status of the Set Objective}
The three terms of $F(\bm{m})$ are surrogates taken from the submodular set-function
literature we build on rather than regularizers assembled by trial and error.
$\text{Rel}(\bm{m})$ is a modular relevance term, $\text{Cov}(\bm{m})$ is a
log-sum-exp coverage surrogate whose curvature is what produces diminishing returns,
and $\text{Red}(\bm{m})$ is a pairwise dispersion penalty. Two of the three,
$\text{Cov}(\bm{m})$ and $\text{Red}(\bm{m})$, couple the frames to one another, so
$F(\bm{m})$ scores a subset as a subset instead of summing independent per-frame scores.
That coupling is what the word holistic refers to throughout the paper, and it is also
what separates the objective from point-wise scoring, which cannot express the fact that
the value of a frame depends on which other frames were already chosen. The weights of
the three terms are not learned, and Table~\ref{tab:submodular-weights} shows that the
objective is not delicately tuned either. That table reports the sweep behind the
claim in Section~\ref{sec:ablation}, with every configuration given per duration split.

\begin{table}[t]
\centering
\caption{Ablation study on the set-level optimization weight configuration on Video-MME
with subtitles, using Qwen2.5-VL-7B-Instruct as the teacher model.}
\label{tab:submodular-weights}
\small
{
\begin{tabular}{ccccccc}
\toprule
\multirow{2}{*}[0.5ex]{\textbf{$\lambda_{\text{rel}}$}} & \multirow{2}{*}[0.5ex]{\textbf{$\lambda_{\text{cov}}$}} & \multirow{2}{*}[0.5ex]{\textbf{$\lambda_{\text{red}}$}} & \multicolumn{4}{c}{\textbf{Video-MME}} \\
\cmidrule(lr){4-7}
& & & \textbf{Short} & \textbf{Medium} & \textbf{Long} & \textbf{Overall} \\
\midrule
0.4 & 0.4 & 0.2 & 72.0 & 59.9 & 54.1 & 62.0 \\
0.5 & 0.2 & 0.3 & 71.9 & 60.1 & 54.3 & 62.1 \\
\rowcolor{lightpurple}
0.5 & 0.3 & 0.2 & \textbf{72.2} & \textbf{60.4} & \textbf{55.0} & \textbf{62.6} \\
0.3 & 0.3 & 0.4 & 71.3 & 59.0 & 53.2 & 61.2 \\
\bottomrule
\end{tabular}
}
\end{table}

\subsection{Teacher Scorer}
\label{app:teacher-scorer}
The mutual learning objective of Section~\ref{sec:mutual} requires the teacher reasoner to expose a frame-importance distribution of its own. This subsection specifies the scorer that produces it, so that the alignment term $\mathcal{L}_{\text{KL}}$ can be reproduced exactly.

\noindent\textbf{From teacher hidden states to per-frame vectors.} The teacher receives the $k_{\text{sel}}$ selected frames as an interleaved sequence, one visual item followed by its timestamp text, and then the question and the options, as laid out in Box 2. We run a single forward pass with hidden states enabled and read the second-to-last layer. Visual tokens are located through the visual placeholder that the processor inserts into the input identifiers, so no assumption is made about where the frame region begins or ends in the sequence. The processor also returns the visual grid for the batch, which gives the number of tokens each frame expands into after patch merging, and we use it to partition the located visual tokens into $k_{\text{sel}}$ groups, one per selected frame. Mean pooling within group $i$ gives $\bm{h}_i \in \mathbb{R}^{d_h}$, and stacking the groups gives $\bm{H}_{\mathcal{M}_t}$. The hidden state at the last non-padding position is taken as the global context vector $\bm{h}_{\text{con}}$, which summarizes the question and the options. Because the grouping is driven entirely by the processor output, it transfers unchanged between the two teachers, whose per-frame token counts differ.

\noindent\textbf{Scorer head.} For each selected frame the pooled teacher representation $\bm{h}_i$ is concatenated with the global context vector $\bm{h}_{\text{con}}$, so that the same frame can be scored differently under different questions. The resulting $2 d_h$-dimensional vector is passed through $\text{MLP}_t$, a three-layer network with widths $2 d_h \rightarrow 1024 \rightarrow 512 \rightarrow 1$, where each of the first two linear layers is followed by layer normalization, a GELU nonlinearity and dropout with rate $0.1$. Both teachers we use have $d_h = 3584$, so $\text{MLP}_t$ holds $7.87$M parameters, which is $37\%$ of the $20.98$M trainable parameters reported in Table~\ref{tab:training-cost}. The head is shared across frames and applied independently to each of them, so it is agnostic to $k_{\text{sel}}$.

\noindent\textbf{Normalization across frames.} The $k_{\text{sel}}$ scalars emitted by $\text{MLP}_t$ are normalized by a softmax taken over the selected frames, which yields $\bm{p}_{\mathcal{M}_t}$. The student distribution $\bm{p}_{\mathcal{M}_s}$ is a softmax over the same index set $S$, applied to the student scores $\{s_i\}_{i \in S}$ divided by the distillation temperature $\tau_d = 0.5$. Both distributions are therefore supported on the same $k_{\text{sel}}$ indices, and $\mathcal{L}_{\text{KL}}$ is well defined without any further alignment step. We normalize over the selected subset rather than over all $N$ candidates on purpose, because the teacher only observes the subset and its judgment is meaningful only relative to what it was shown.

\noindent\textbf{Gradient flow and cost.} The teacher hidden states are retained in the autograd graph, and the KL target is not detached, so gradients from $\mathcal{L}_{\text{KL}}$ reach $\text{MLP}_t$ and the teacher LoRA adapter as well as the student scorer. This is what makes the alignment mutual rather than a one-directional distillation from a frozen target, and it is also why $\lambda_{\text{KL}}$ is warmed up from $0.1$ to $1.0$ over the first epoch, since an untrained $\text{MLP}_t$ would otherwise pull the student toward a near-uniform distribution. The scorer participates in training only. At inference the student produces $S$, the teacher answers from the selected frames, and no importance distribution is computed, so $\text{MLP}_t$ contributes nothing to the latencies in Table~\ref{tab:latency}.

\section{Additional Quantitative Analyses}
\label{app:additional}
This appendix reports the supporting analyses referenced from the main text, grouped by the question each one answers.

\subsection{Training Cost}
Table~\ref{tab:training-cost} compares the one-time training cost of HFS against a control that fine-tunes only the teacher LoRA adapter, with the data, schedule and batch size held fixed. HFS updates 20.98M parameters against 2.52M, raises peak memory from 24.9G to 34.4G, and costs 0.766s per step against 0.649s, an increase of 18\% per step. HFS is trained for a single epoch of about 40 GPU-h on one H100, so the 18\% overhead is paid once, before deployment.

\begin{table}[t]
\centering
\caption{Training cost of HFS versus fine-tuning only the teacher LoRA baseline, on a single H100 with the Qwen2.5-VL-7B teacher. The overhead is incurred once, during a single training epoch of about 40 GPU-h.}
\label{tab:training-cost}
\small
{
\begin{tabular}{lcc}
\toprule
\textbf{Metric} & \textbf{Teacher-LoRA} & \textbf{HFS} \\
\midrule
Trainable params & 2.52M & 20.98M \\
Peak VRAM & 24.9G & 34.4G \\
Time / step & 0.649s & 0.766s \\
\bottomrule
\end{tabular}
}
\end{table}

\subsection{Training Stability and Hyperparameter Sensitivity}
Figure~\ref{fig:training-loss} plots the total training loss over the full epoch. The curve decreases smoothly and without divergence, which confirms that the Gumbel-TopK relaxation, combined with temperature annealing, trains stably in our setting.

Table~\ref{tab:tau-ablation} ablates the temperature schedule itself. Annealing $\tau$ from 2.0 to 0.5 reaches 63.3 on Video-MME, ahead of every fixed temperature, the best of which is 62.8 at $\tau=1.0$. Early exploration followed by late commitment therefore matters more than any single operating point.

Table~\ref{tab:gamma-sweep} varies the temporal-kernel bandwidth $\gamma$ and reports each duration split separately, since a bandwidth defined on absolute time would favor one split over another. No split moves by more than 1.1 points across the whole sweep, and $\gamma=10$ is best overall. This follows from the definition of $t_i$ given with the kernel in the method: because the kernel operates on the normalized index grid of the $N$ uniformly sampled candidates, $\gamma$ is independent of absolute video duration and a single value transfers across short, medium and long videos.

\begin{figure}[t]
  \centering
  \includegraphics[width=\columnwidth]{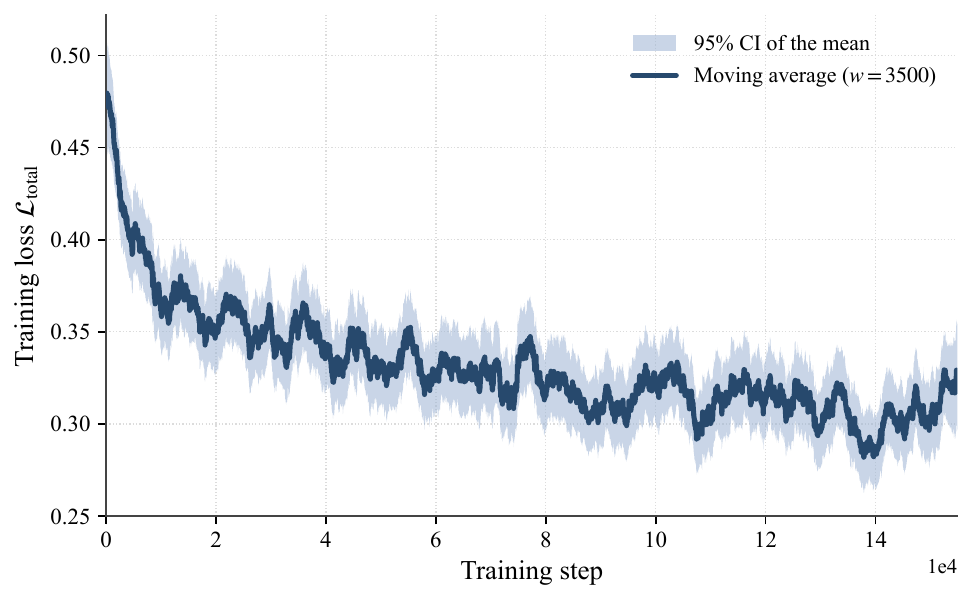}
  \Description{A line plot of the total training loss decreasing smoothly over one epoch of training.}
  \caption{Training loss $\mathcal{L}_{\text{total}}$ over the full epoch. The loss decreases smoothly, confirming that Gumbel-TopK training is stable in our setting. One epoch takes about 40 GPU-h on a single H100.}
  \label{fig:training-loss}
\end{figure}

\begin{table}[t]
\centering
\caption{Ablation on the Gumbel-TopK temperature schedule $\tau$, on Video-MME (without subtitles, InternVL3-8B teacher). The annealed schedule is best. \colorbox{lightpurple}{Highlighted row} is our final configuration.}
\label{tab:tau-ablation}
\small
{
\begin{tabular}{lc}
\toprule
\textbf{$\tau$ schedule} & \textbf{Video-MME (w/o sub.)} \\
\midrule
Fixed $0.5$ & 62.4 \\
Fixed $1.0$ & 62.8 \\
Fixed $2.0$ & 62.7 \\
\rowcolor{lightpurple}
Annealed $2.0\!\to\!0.5$ & \textbf{63.3} \\
\bottomrule
\end{tabular}
}
\end{table}

\begin{table}[t]
\centering
\caption{Sensitivity to the temporal-kernel bandwidth $\gamma$, by video duration, on Video-MME (without subtitles, InternVL3-8B teacher). A fixed $\gamma{=}10$ is robust across all durations. \colorbox{lightpurple}{Highlighted row} is our final configuration.}
\label{tab:gamma-sweep}
\small
{
\begin{tabular}{lcccc}
\toprule
\textbf{$\gamma$} & \textbf{Short} & \textbf{Medium} & \textbf{Long} & \textbf{Overall} \\
\midrule
5 & 73.9 & 59.6 & 55.8 & 63.1 \\
\rowcolor{lightpurple}
10 & 73.8 & 59.8 & 56.4 & \textbf{63.3} \\
20 & 73.2 & 59.2 & 56.7 & 63.0 \\
50 & 72.9 & 59.4 & 55.6 & 62.6 \\
\bottomrule
\end{tabular}
}
\end{table}

\subsection{Statistical Reliability}
Table~\ref{tab:bootstrap-ci} reports bootstrap 95\% confidence intervals for HFS on each benchmark, obtained by resampling the corresponding evaluation split. The width tracks split size, from $\pm0.8$ on NExT-QA to $\pm4.5$ on MLVU. These are marginal intervals for the absolute accuracy of a single system. Because every selector in our comparison is evaluated on exactly the same questions with the same teacher and the same frame budget, the paired difference between two systems carries considerably less variance than a comparison of two marginal intervals would suggest.

\begin{table}[t]
\centering
\caption{Bootstrap 95\% confidence intervals of HFS (InternVL3-8B teacher) on the four benchmarks, obtained by resampling the corresponding evaluation split.}
\label{tab:bootstrap-ci}
\small
{
\begin{tabular}{lc}
\toprule
\textbf{Benchmark} & \textbf{Accuracy (95\% CI)} \\
\midrule
Video-MME (w/ sub.) & $66.1 \pm 1.9$ \\
LongVideoBench & $60.2 \pm 2.7$ \\
MLVU & $50.0 \pm 4.5$ \\
NExT-QA & $83.1 \pm 0.8$ \\
\bottomrule
\end{tabular}
}
\end{table}

\subsection{Query Diversity and Selection Redundancy}
Table~\ref{tab:lsep-diversity} isolates the effect of the separation loss, averaged over the three pairs formed by the $K{=}3$ latent queries. Without $\mathcal{L}_{\text{sep}}$ the three queries behave almost identically, sharing 12.84 of their top-16 frames and correlating at 0.85 in per-frame importance. Adding $\mathcal{L}_{\text{sep}}$ nearly halves the overlap to 6.65, lowers the rank correlation to 0.56, and drives the squared cosine similarity between query vectors from 0.59 down to 0.12. Figure~\ref{fig:attention-heatmap} shows the same effect directly: the three queries attend to complementary regions of the candidate sequence rather than collapsing onto one intent.

Table~\ref{tab:interframe-sim} then measures the redundancy of the final selection with a parameter-free metric, the average pairwise CLIP cosine similarity among the 16 selected frames. HFS attains the lowest value on both splits, 0.672 and 0.683, below the diversity-specialized \text{\small M}DP$^3$ at 0.680 and 0.702, while remaining query-aware and 2.4 times faster at inference.

\begin{table}[t]
\centering
\caption{Effect of the separation loss $\mathcal{L}_{\text{sep}}$ on cross-query frame-selection diversity, averaged over the three pairs of the $K{=}3$ latent queries on Video-MME. Lower is more diverse: overlap@16 counts shared frames in the top-16 sets, Spear.\ is the Spearman correlation of per-frame importance, and $\cos^2$ is the squared cosine similarity between query vectors. \colorbox{lightpurple}{Highlighted row} is our final configuration.}
\label{tab:lsep-diversity}
\small
{
\begin{tabular}{lccc}
\toprule
\textbf{Setting} & \textbf{overlap@16}$\downarrow$ & \textbf{Spear.}$\downarrow$ & \textbf{$\cos^2$}$\downarrow$ \\
\midrule
w/o $\mathcal{L}_{\text{sep}}$ & 12.84 & 0.85 & 0.59 \\
\rowcolor{lightpurple}
HFS (w/ $\mathcal{L}_{\text{sep}}$) & \textbf{6.65} & \textbf{0.56} & \textbf{0.12} \\
\bottomrule
\end{tabular}
}
\end{table}

\begin{table}[t]
\centering
\caption{Inter-frame similarity of the 16 selected keyframes (average pairwise CLIP cosine similarity) on the MLVU test and LongVideoBench validation splits. Lower is less redundant; HFS is the least redundant while remaining query-aware.}
\label{tab:interframe-sim}
\small
{
\begin{tabular}{lcc}
\toprule
\textbf{Method} & \textbf{MLVU}$\downarrow$ & \textbf{LongVideoBench}$\downarrow$ \\
\midrule
Uniform & 0.715 & 0.698 \\
TopK+Threshold & 0.696 & 0.698 \\
AKS & 0.734 & 0.784 \\
Q-Frame & 0.722 & 0.716 \\
\text{\small M}DP$^3$ & 0.680 & 0.702 \\
\rowcolor{lightpurple}
HFS & \textbf{0.672} & \textbf{0.683} \\
\bottomrule
\end{tabular}
}
\end{table}

\begin{figure}[t]
  \centering
  \includegraphics[width=\columnwidth]{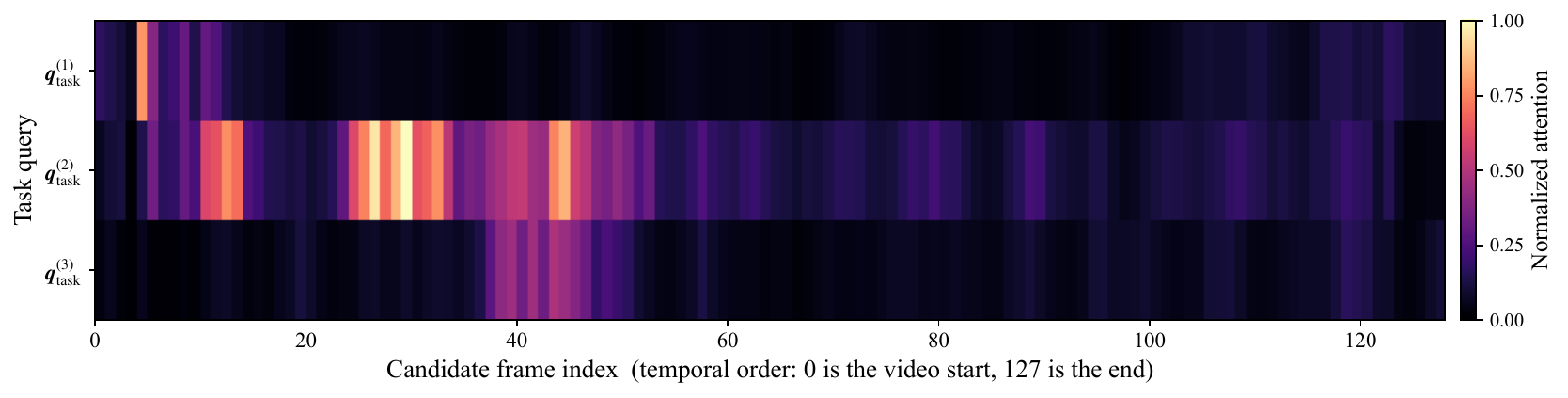}
  \Description{Attention heatmaps for three latent queries over 128 candidate frames, showing that the three queries attend to different frames.}
  \caption{Per-query attention over the 128 candidate frames for the $K{=}3$ latent queries, drawn from a representative video and its question. The three queries attend to complementary frames, indicating that they capture distinct semantic facets rather than collapsing to a single intent.}
  \label{fig:attention-heatmap}
\end{figure}

\subsection{Robustness to Query Paraphrasing}
Because the latent queries are extracted from the question text, a natural concern is whether HFS is sensitive to how a question happens to be worded. Table~\ref{tab:paraphrase} tests this directly. To avoid any circular dependence on the models under test, we use a separate paraphraser, Llama-3.3-70B-Instruct, to rewrite all 2K NExT-QA test questions while preserving intent, applying one of three transformations to each question: word-order reshuffling, synonym substitution, or active-passive conversion. Accuracy moves by 0.2 points under Qwen2.5-VL and 0.6 points under InternVL3, well inside the intervals of Table~\ref{tab:bootstrap-ci}, indicating that the extracted queries track task intent rather than surface wording.

\begin{table}[t]
\centering
\caption{Robustness to query paraphrasing on a 2K subset of the NExT-QA test set. Questions are rewritten by Llama-3.3-70B-Instruct (word-order reshuffling, synonym substitution, or active-passive conversion) while preserving intent. Accuracy is essentially unchanged.}
\label{tab:paraphrase}
\small
{
\begin{tabular}{lcc}
\toprule
\textbf{Setting} & \textbf{Qwen2.5-VL} & \textbf{InternVL3} \\
\midrule
Original questions & 78.6 & 81.9 \\
Paraphrased questions & 78.4 & 81.3 \\
\bottomrule
\end{tabular}
}
\end{table}

\subsection{Initial Candidate Pool Size}
\begin{table}[t]
\centering
\caption{Effect of the initial candidate pool size $N$ on Video-MME with subtitles and LongVideoBench (Qwen2.5-VL teacher), when selecting $k_{\text{sel}}{=}16$ frames; each $N$ is retrained. $N{=}128$ is the best-balanced default. \colorbox{lightpurple}{Highlighted row} is our final configuration.}
\label{tab:n-sweep}
\small
{
\begin{tabular}{lcc}
\toprule
\textbf{$N$} & \textbf{Video-MME} & \textbf{LongVideoBench} \\
\midrule
64 & 61.4 & 56.2 \\
\rowcolor{lightpurple}
128 & \textbf{62.6} & \textbf{57.3} \\
256 & 62.0 & 56.9 \\
\bottomrule
\end{tabular}
}
\end{table}
\begin{table*}[!t]
\centering
\small
\setlength{\tabcolsep}{4.8pt}
\caption{Detailed results on the validation set of LongVideoBench benchmark. \textcolor{red}{Red} values indicate improvements over the baseline, while \textcolor{blue}{blue} values indicate degradation.}
\label{tab:longvideobench}

\begin{tabular}{lccccccccccccccccc}
\toprule
\multirow{2}{*}{\textbf{Model}} & \multicolumn{17}{c}{\textbf{LongVideoBench}} \\
\cmidrule{2-18}
& \textbf{E2O} & \textbf{E3E} & \textbf{O2E} & \textbf{O3O} & \textbf{S2A} & \textbf{S2E} & \textbf{S2O} & \textbf{SAA} & \textbf{SOS} & \textbf{SSS} & \textbf{T2A} & \textbf{T2E} & \textbf{T2O} & \textbf{T3E} & \textbf{T3O} & \textbf{TAA} & \textbf{TOS} \\
\midrule
Qwen2.5-VL~\cite{bai2025qwen25vl} & 59.4 & 63.8 & 63.2 & 48.5 & 71.6 & 65.6 & 59.7 & 47.2 & 53.1 & 33.0 & 60.8 & 67.7 & 44.7 & 53.4 & 45.9 & 47.6 & 40.5 \\
Qwen2.5-VL~\cite{bai2025qwen25vl} + \textbf{HFS} & \textcolor{red}{62.5} & \textcolor{red}{66.0} & \textcolor{red}{66.7} & \textcolor{red}{51.5} & 71.6 & \textcolor{red}{67.7} & \textcolor{red}{63.9} & \textcolor{red}{48.6} & \textcolor{red}{59.3} & \textcolor{red}{34.0} & \textcolor{blue}{59.5} & \textcolor{red}{70.8} & \textcolor{red}{50.0} & \textcolor{red}{56.2} & \textcolor{red}{50.0} & \textcolor{red}{51.2} & \textcolor{red}{44.6} \\
\midrule
InternVL3~\cite{zhu2025internvl3} & 60.9 & 66.0 & 65.5 & 51.5 & 73.9 & 68.8 & 62.5 & 50.0 & 54.3 & 36.1 & 63.3 & 69.2 & 46.1 & 54.8 & 47.3 & 48.8 & 43.2 \\
InternVL3~\cite{zhu2025internvl3} + \textbf{HFS} & \textcolor{red}{70.3} & \textcolor{blue}{63.8} & \textcolor{red}{73.6} & \textcolor{red}{54.5} & \textcolor{blue}{71.6} & \textcolor{red}{69.9} & \textcolor{red}{66.7} & \textcolor{blue}{48.6} & \textcolor{red}{67.9} & \textcolor{red}{37.1} & \textcolor{blue}{62.0} & \textcolor{red}{70.8} & \textcolor{red}{56.6} & \textcolor{blue}{53.4} & \textcolor{red}{63.5} & \textcolor{red}{50.0} & \textcolor{red}{44.6} \\
\bottomrule
\end{tabular}

\end{table*}
\begin{table}[!t]
\centering
\caption{Detailed results on the test set of NExT-QA benchmark. Both backbones improve on all three question types.}
\label{tab:nqa}
\small
{
\begin{tabular}{lccc}
\toprule
\multirow{2}{*}{\textbf{Model}} & \multicolumn{3}{c}{\textbf{NExT-QA}} \\
\cmidrule{2-4}
& \textbf{Causal} & \textbf{Temporal} & \textbf{Descriptive} \\
\midrule
Qwen2.5-VL~\cite{bai2025qwen25vl} & 77.6 & 76.5 & 80.9\\
Qwen2.5-VL~\cite{bai2025qwen25vl} + \textbf{HFS} & \textcolor{red}{79.3} & \textcolor{red}{76.8} & \textcolor{red}{85.4}\\
\midrule
InternVL3~\cite{zhu2025internvl3} & 81.5 & 79.5 & 87.6 \\
InternVL3~\cite{zhu2025internvl3} + \textbf{HFS} & \textcolor{red}{82.1} & \textcolor{red}{81.6} & \textcolor{red}{89.1}\\
\bottomrule
\end{tabular}
}
\end{table}

HFS selects $k_{\text{sel}}$ frames from a pool of $N$ uniformly sampled candidates, so $N$ bounds what the selector can possibly recover. Table~\ref{tab:n-sweep} retrains the model separately for each $N$. Setting $N{=}64$ costs 1.2 and 1.1 points on the two benchmarks, since too few candidates already miss the decisive frames, whereas $N{=}256$ yields no further gain even after retraining. The binding constraint is therefore the downstream frame budget $k_{\text{sel}}{=}16$ rather than candidate density. Appendix~\ref{app:limitations} discusses what this implies for events shorter than the inter-candidate spacing.

\section{Detailed Benchmark Results}
\label{app:benchmark_results}
We present results per category on two benchmarks. For Long\-Video\-Bench~\cite{Wu_2024_NeurIPS}, we report results on 17 referring reasoning categories as defined in the original paper. These categories are divided into two levels. Perception tasks include S2E (Scene-referred Event), S2O (Scene-referred Object Existence), S2A (Scene-referred Object Attribute), E2O (Event-referred Object), O2E (Object-referred Event), T2E (Text-referred Event), T2O (Text-referred Object Existence), and T2A (Text-referred Object Attribute). Relation tasks include E3E (Event before/after Event), O3O (Object before/after Object), SSS (Sequence of Scenes), SOS (Scene-referred Object Tracking), SAA (Scene-referred Object Attribute Change), T3E (Event before/after Text), T3O (Object before/after Text), TOS (Text-referred Object Tracking), and TAA (Text-referred Object Attribute Change). For the test set of NExT-QA~\cite{Xiao_2021_CVPR}, we report on three question types: causal, temporal, and descriptive. In both tables a \textcolor{red}{red} value marks a category on which HFS improves over the corresponding baseline and a \textcolor{blue}{blue} value one on which it declines, and a black value an unchanged score.

The detailed results on LongVideoBench validation set in Table~\ref{tab:longvideobench} demonstrate that HFS improves performance across many categories for both baseline models. For Qwen2.5-VL~\cite{bai2025qwen25vl}, HFS achieves improvements in 15 out of 17 categories. The only marginal decline occurs in T2A, which is negligible compared to the overall performance boost. The pattern is consistent with what the selector is trained to do: the categories that gain most are the ones whose evidence sits in a small number of frames, so choosing those frames is most of the task, whereas a category whose answer needs a fine attribute reading depends on what the reasoner can resolve once the frames are in front of it. 

When combined with InternVL3~\cite{zhu2025internvl3}, HFS shows pronounced improvements on reasoning tasks, achieving remarkable gains. However, we observe slight performance drops in five categories, primarily those involving fine-grained attribute recognition. This suggests that while HFS excels at identifying key frames for object localization and event understanding, there remains room for improvement in tasks requiring detailed attribute-level discrimination.

The results on the NExT-QA test set reveal consistent improvements across all question types for both models, as shown in Table~\ref{tab:nqa}. Qwen2.5-VL~\cite{bai2025qwen25vl} with HFS achieves the most significant gain in descriptive questions, while showing steady improvements in causal and temporal reasoning. InternVL3~\cite{zhu2025internvl3} benefits from HFS in temporal reasoning tasks, alongside improvements in causal and descriptive questions.

Across both benchmarks, HFS consistently demonstrates that selecting a query-aware subset of frames yields superior performance compared to uniform sampling. HFS effectively mitigates the information loss often associated with fixed-interval sampling. These quantitative results support our hypothesis that holistic query-awareness is essential for efficient video reasoning.

\begin{figure}[!t]
  \centering
  \includegraphics[width=\columnwidth]{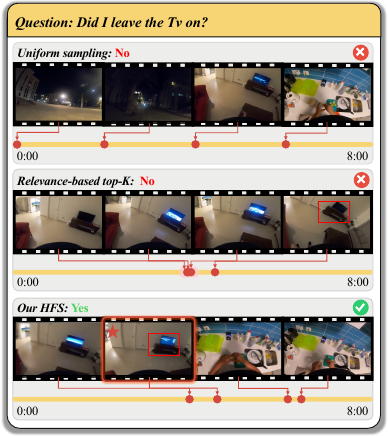}
  \Description{Three rows of video frames comparing uniform sampling, top-K selection, and HFS selection, with the HFS row showing a TV in the on state.}
  \caption{Qualitative comparison on an egocentric reasoning task. Uniform sampling spends most of its budget on outdoor and kitchen scenes and answers wrongly. Relevance-based top-K sampling successfully localizes the ``TV'' object; however, the final selected frame depicts the ``TV off'' state. In contrast, our HFS also localizes the ``TV'' object, and the frame it highlights shows the ``TV on'' state.}
  \label{fig:v}
\end{figure}

\section{Qualitative Analysis}
\label{app:qualitative_analysis}
Figure~\ref{fig:v} shows the failure mode that motivates query-aware selection. The task is to identify the running status of a television in a long egocentric video. Uniform sampling spends most of its budget on scenes without the television. Relevance-based top-K sampling locates the television but shows it in the off state, because shallow visual similarity cannot distinguish task-relevant state changes. HFS instead uses the CoT-conditioned queries to infer that the operational status is what matters, and selects a frame that correctly shows the television on. Both baselines would be scored as attending to the right object, yet only one of them supports the right answer, which is precisely the distinction that aggregate accuracy hides.

We then present five further examples in Figure~\ref{fig:qualitative_vis}, Figure~\ref{fig:qualitative_vis2}, Figure~\ref{fig:qualitative_vis3}, Figure~\ref{fig:qualitative_vis4}, and Figure~\ref{fig:qualitative_vis5}. In each of them, we display all 128 candidate frames sampled from the video as a grid, and overlay three sets of colored markers to indicate which 16 frames are selected by each method. The question, the candidate options, and the ground-truth answer, highlighted in red, are shown above the frame grid. The five are chosen so that the demand each one places on selection is different. Figure~\ref{fig:qualitative_vis} asks what a cartoon character conjures with a wand, so the answer exists in a single instant and a selector that spreads its budget evenly is likely to step over it. Figure~\ref{fig:qualitative_vis2} asks what type of house is under construction, which is the opposite situation, because the evidence accumulates over the whole video and no individual frame settles the question. Figure~\ref{fig:qualitative_vis3} is an action recognition case in which many frames are partially informative and the difficulty is redundancy rather than scarcity. Figure~\ref{fig:qualitative_vis4} asks about a collision in a rendered scene, so the relevant interval is short and surrounded by visually similar frames. Figure~\ref{fig:qualitative_vis5} asks where an object was before the action named in the question, so the frames that matter lie before the moment a query-agnostic method would consider salient. Taken together they span question types from object localization and action recognition to causal reasoning over rare events, and video sources from synthetic renderings to unedited first-person capture. Together, they illustrate how HFS adapts its selection pattern to the specific intent of each question, and how the resulting frame subset differs qualitatively from those produced by uniform sampling and relevance-based top-K baselines.

\section{Limitations and Future Work} \label{app:limitations} While HFS achieves consistent improvements over existing frame selection methods across four video understanding benchmarks, we acknowledge a key limitation that points to a promising direction for future research. The latent query vectors $\{\bm{q}_\text{task}^{(k)}\}_{k=1}^K$ are extracted from hidden states that the student model produces when conditioned on the question and its candidate options. When the question is ambiguous, or contains minimal semantic content (e.g., generic questions such as ``What happens in the video?''), the resulting queries may fail to capture a sufficiently discriminative task intent.

A second consideration is the fixed initial candidate pool of $N{=}128$ uniformly sampled frames. Because the pool is sampled uniformly, events shorter than the inter-candidate spacing may be missed before selection begins. In practice, $N{=}128$ is a well-balanced default rather than a bottleneck. When retrained on Qwen2.5-VL with $N \in \{64, 128, 256\}$ (Table~\ref{tab:n-sweep}), $N{=}128$ is best, $N{=}64$ drops because too few candidates miss key frames, and $N{=}256$ yields no further gain even after retraining. The binding constraint is therefore the downstream frame budget $k_{\text{sel}}{=}16$ rather than candidate density, and adaptive candidate generation is a complementary direction for future work.

\begin{figure*}[p]
    \centering
    \includegraphics[width=\textwidth,height=0.9\textheight,keepaspectratio]{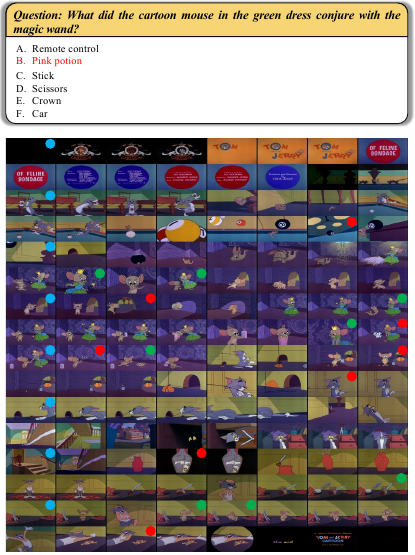}
    \Description{A grid of 128 candidate video frames from an animated cartoon, overlaid with blue, green and red markers showing which sixteen frames uniform sampling, relevance-based top-K and HFS each select.}
    \caption{Qualitative comparison of frame selection methods on a video question answering example. The blue dots indicate frames selected by uniform sampling, the green dots indicate frames selected by the relevance-based top-K method, and the red dots indicate frames selected by our HFS method.}
    \label{fig:qualitative_vis}
\end{figure*}

\begin{figure*}[p]
    \centering
    \includegraphics[width=\textwidth,height=0.9\textheight,keepaspectratio]{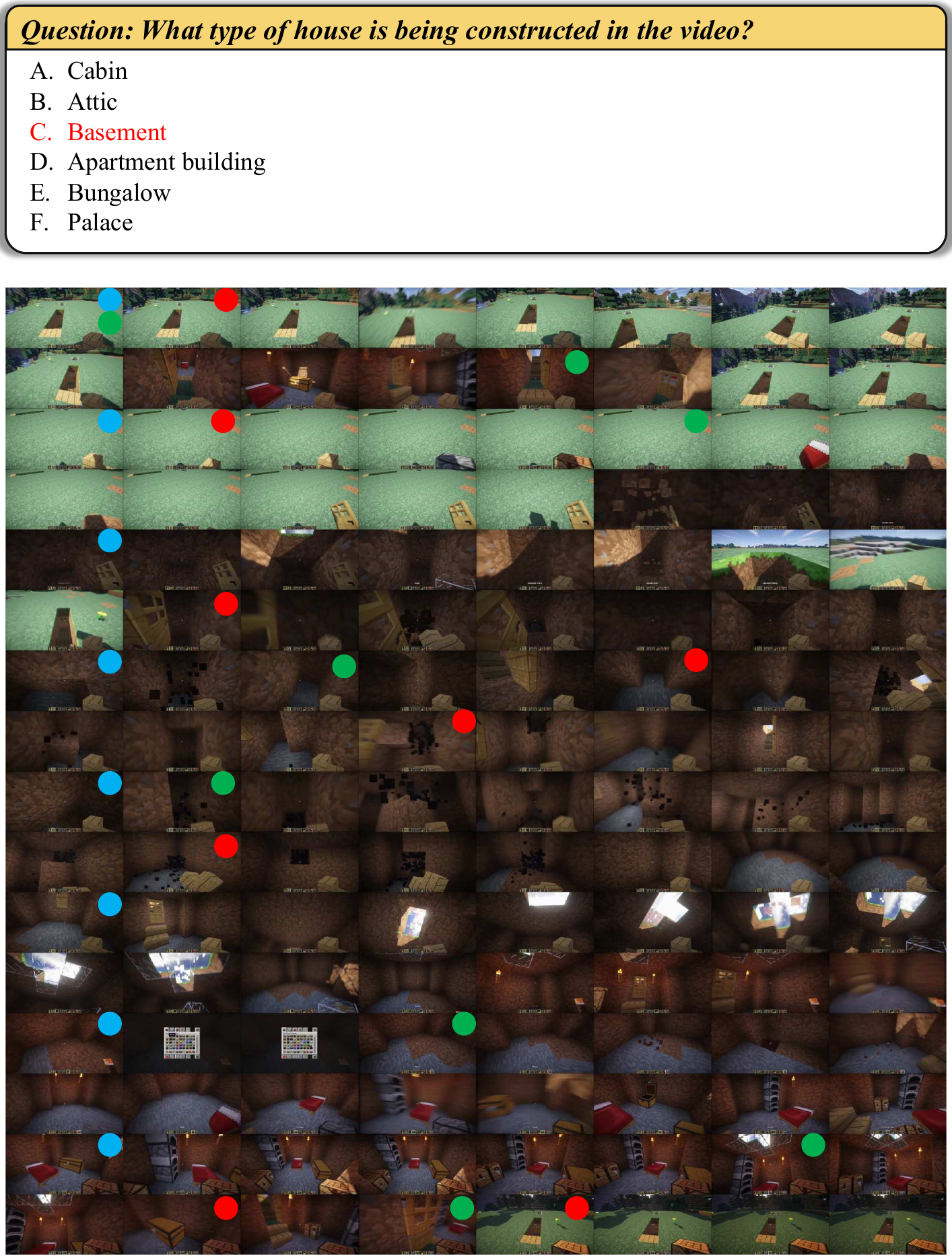}
    \Description{A grid of 128 candidate video frames from a house construction video, overlaid with blue, green and red markers showing which sixteen frames each selection method chooses.}
    \caption{Qualitative comparison of frame selection methods on a video question answering example.}
    \label{fig:qualitative_vis2}
\end{figure*}

\begin{figure*}[p]
    \centering
    \includegraphics[width=\textwidth,height=0.9\textheight,keepaspectratio]{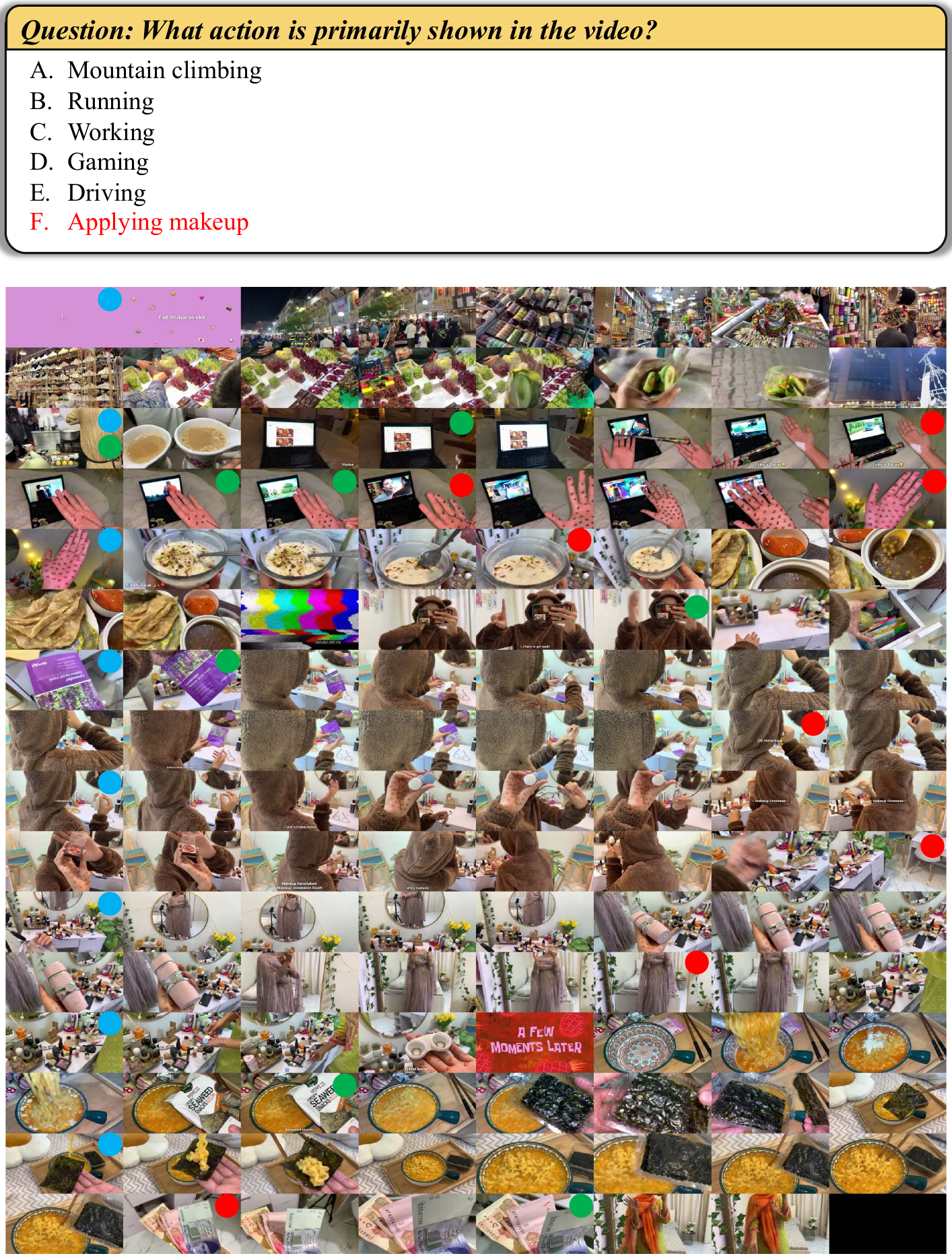}
    \Description{A grid of 128 candidate video frames from an action recognition example, overlaid with blue, green and red markers showing which sixteen frames each selection method chooses.}
    \caption{Qualitative comparison of frame selection methods on a video question answering example.}
    \label{fig:qualitative_vis3}
\end{figure*}

\begin{figure*}[p]
    \centering
    \includegraphics[width=\textwidth,height=0.9\textheight,keepaspectratio]{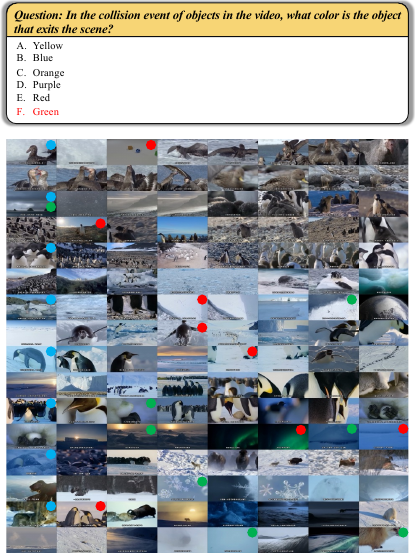}
    \Description{A grid of 128 candidate video frames from a rendered object collision scene, overlaid with blue, green and red markers showing which sixteen frames each selection method chooses.}
    \caption{Qualitative comparison of frame selection methods on a video question answering example.}
    \label{fig:qualitative_vis4}
\end{figure*}

\begin{figure*}[p]
    \centering \includegraphics[width=\textwidth,height=0.9\textheight,keepaspectratio]{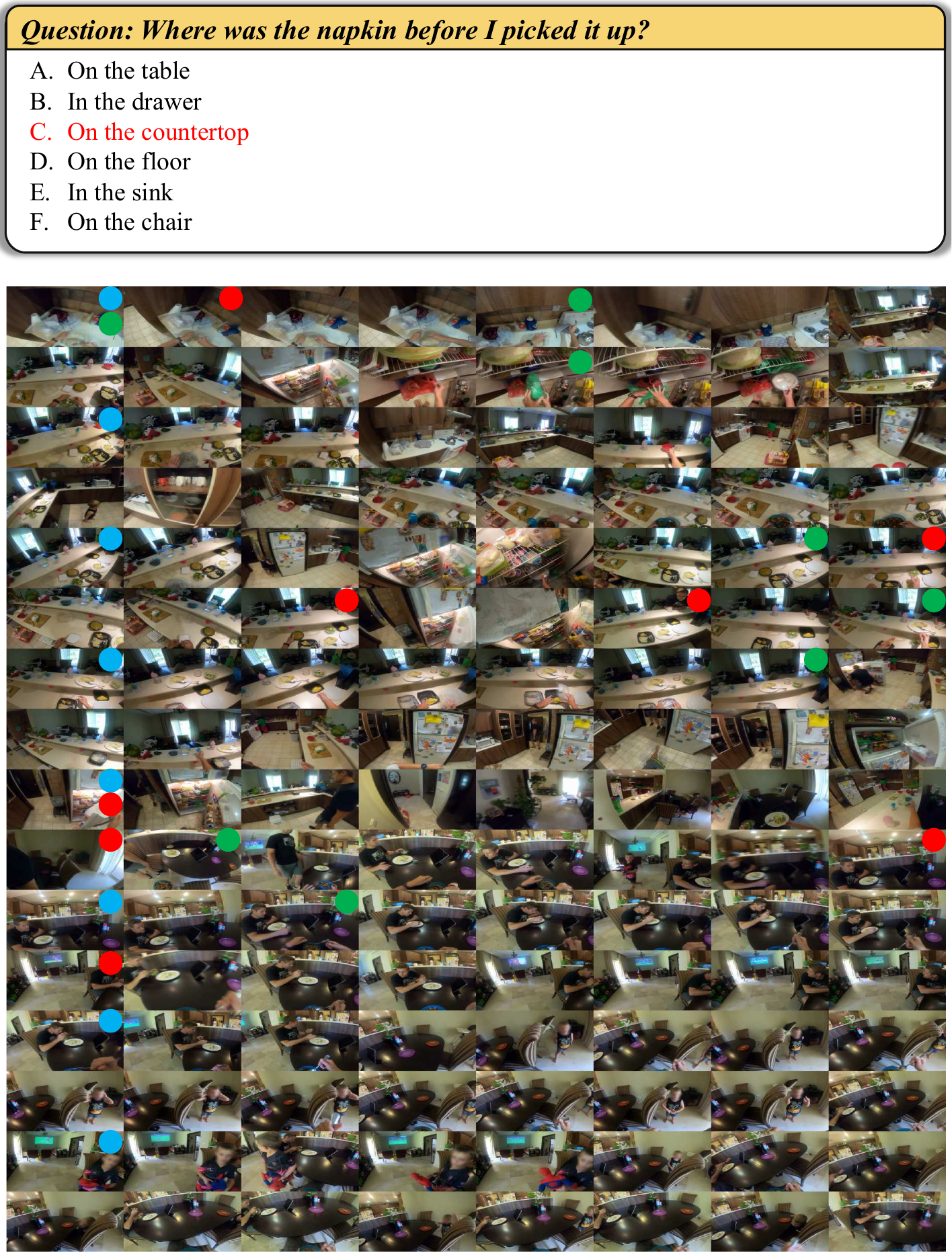}
    \Description{A grid of 128 candidate video frames from egocentric kitchen footage, overlaid with blue, green and red markers showing which sixteen frames each selection method chooses.}
    \caption{Qualitative comparison of frame selection methods on a video question answering example.}
    \label{fig:qualitative_vis5}
\end{figure*}
\fi

\end{document}